%% file: main_icml_workshop.tex
\documentclass{article}

\usepackage{cmap}
\usepackage[T1]{fontenc}

\usepackage{microtype}
\usepackage{graphicx}
\usepackage{subcaption}
\usepackage{booktabs}
\PassOptionsToPackage{hyphens}{url}  % allow URLs to break at hyphens/slashes (xurl unavailable in this TeX dist)
\usepackage{hyperref}

\usepackage[preprint]{icml2026}

\makeatletter
\renewcommand{\ICML@preprint}{\textit{Accepted at the ICML 2026 Workshop on Human-AI Co-Creativity (non-archival).}}
\makeatother

\usepackage{amsmath}
\usepackage{amssymb}
\usepackage{amsthm}
\usepackage{enumitem}

\graphicspath{{figures/}{investigations/cross_mode_surprise_drop/figures/}}
\usepackage{pgfplots}
\usepgfplotslibrary{fillbetween}
\pgfplotsset{compat=1.18}
\usetikzlibrary{positioning, arrows.meta, fit, backgrounds, calc, shapes.geometric}

\newcommand{\E}{\mathbb{E}}

\newcommand{\TC}{\mathrm{TC}}

\newif\ifworkshop \workshoptrue

\newcommand{\projectGithubUrl}{\url{https://github.com/AMindToThink/icl-diversity}}

\input{results/tables/paper_macros.tex}

\icmltitlerunning{\textquotedblleft I've Seen How This Goes\textquotedblright: Characterizing Diversity via Progressive Conditional Surprise}

\begin{document}

\twocolumn[
\icmltitle{\textquotedblleft I've Seen How This Goes\textquotedblright: Characterizing the Diversity of LLM Generations and Human Writing via Progressive Conditional Surprise}

\icmlsetsymbol{equal}{*}

\begin{icmlauthorlist}
\icmlauthor{Matthew Khoriaty}{era}
\icmlauthor{David Williams-King}{era}
\icmlauthor{Shi Feng}{gwu}
\end{icmlauthorlist}

\icmlaffiliation{era}{ERA Fellowship}
\icmlaffiliation{gwu}{George Washington University}

\icmlcorrespondingauthor{Matthew Khoriaty}{matthewkhoriaty@gmail.com}

\icmlkeywords{diversity, language models, mode collapse, in-context learning, evaluation, creativity}

\vskip 0.3in
]

\printAffiliationsAndNotice{}

\begin{abstract}
\input{sections/abstract_workshop}
\end{abstract}

% --- Main body (target 4-8 pages; trimmed via Phase 3 workshop variants) ---
% Phase 3 will swap each shared-section \input{} below for a *_workshop.tex
% variant per the plan in paper/human-ai-creativity-workshop.md.

\input{sections/01_motivation_workshop}

\input{sections/02_setup}

\input{sections/03_method_workshop}

\input{sections/10_related}
\input{sections/07_5_tevet_workshop}

\input{sections/07_6_rlhf_workshop}
\input{sections/08_limitations_workshop}
\input{sections/conclusion_workshop}

% Acknowledgements deferred to arXiv v2 (pending contributor consent).
% Text kept in sections/acknowledgements_workshop.tex but intentionally NOT
% \input here for v1; re-add the \input line below to include it in v2.

\input{sections/impact_workshop}

\bibliography{refs}
\bibliographystyle{icml2026}

\newpage
\appendix
% Material trimmed from the 4--8 page main body lives here. Reviewers may skip.
\input{sections/06_practical}
\input{sections/07_experiments_intro}

\input{sections/07_2_scenario_validation}

\input{sections/07_3_mode_count}

\input{sections/07_4_cross_mode}

\input{sections/07_7_practical_findings}

\input{sections/07_8_discussion}

\input{sections/09_future_work}

\input{sections/appB_excess_entropy}

\input{sections/appC_mcdiv_confounds}

\input{sections/appD_aggregation}
\input{sections/appE_qwen3_comparison}
\input{sections/appF_rlhf_cross_metric}

\end{document}

%% file: results/tables/paper_macros.tex
\newcommand{\crossmodeAsymmetryMean}{5.6}
\newcommand{\crossmodeAsymmetryRatio}{3.0}

\newcommand{\crossmodeGPTCrossDamageMten}{-3.4}
\newcommand{\crossmodeGPTDiagGainMten}{8.3}
\newcommand{\crossmodeGPTDiagMean}{83.0}
\newcommand{\crossmodeGPTDiagStd}{20.0}
\newcommand{\crossmodeGPTFirstStepDrop}{+4.9}
\newcommand{\crossmodeGPTNegColCount}{6}
\newcommand{\crossmodeGPTOffMean}{-3.7}
\newcommand{\crossmodeGPTOffPctPos}{30}

\newcommand{\crossmodeNumModes}{15}
\newcommand{\crossmodeNumOffPairs}{210}

\newcommand{\crossmodeQwenDiagMean}{60.5}
\newcommand{\crossmodeQwenDiagStd}{14.4}

\newcommand{\crossmodeQwenNegColCount}{1}
\newcommand{\crossmodeQwenOffMean}{+1.9}
\newcommand{\crossmodeQwenOffPctPos}{64}
\newcommand{\crossmodeQwenOffStd}{2.5}

\newcommand{\scalingFactorialProbPct}{4.2}

\newcommand{\scalingFracPosLlamaEightB}{44}
\newcommand{\scalingFracPosLlamaOneB}{25}
\newcommand{\scalingFracPosLlamaSeventyB}{63}
\newcommand{\scalingFracPosLlamaThreeB}{41}

\newcommand{\scalingNumLlamaModels}{4}

\newcommand{\scalingOffDiagLlamaEightB}{-1.2}
\newcommand{\scalingOffDiagLlamaOneB}{-4.9}
\newcommand{\scalingOffDiagLlamaSeventyB}{+2.1}
\newcommand{\scalingOffDiagLlamaThreeB}{-1.4}

\newcommand{\scalingRCRsqGPT}{0.08}

\newcommand{\scalingRCRsqLlamaSeventyB}{0.09}
\newcommand{\scalingRCRsqLlamaThreeB}{0.67}
\newcommand{\scalingRCRsqQwen}{0.34}
\newcommand{\scalingRCSlopeGPT}{0.35}

\newcommand{\scalingRCSlopeQwen}{0.84}

\newcommand{\scalingSymRsqQwen}{0.24}

\newcommand{\scalingSymSlopeQwen}{0.46}

\newcommand{\qwenThreeBinaryMeanDeltaAUC}{-0.013}

\newcommand{\qwenThreeBinaryTotal}{12}

\newcommand{\qwenThreeBinaryWinsQwenTwoFive}{11}
\newcommand{\qwenThreeConTestPromptGenDeltaAUC}{+0.005}

\newcommand{\qwenThreeDecTestMeanDeltaRho}{+0.021}

\newcommand{\qwenThreeDecTestTotal}{6}
\newcommand{\qwenThreeDecTestWins}{5}

\newcommand{\tevetCoherenceBpbHigh}{2.20}
\newcommand{\tevetCoherenceBpbLow}{0.65}
\newcommand{\tevetCoherenceCHigh}{0.64}
\newcommand{\tevetCoherenceCLow}{0.22}
\newcommand{\tevetCoherenceMeanC}{0.42}
\newcommand{\tevetCoherenceRespCount}{48705}
\newcommand{\tevetCoherenceSetCount}{9741}

\newcommand{\tevetConTestSetCount}{670}
\newcommand{\tevetCxAnVsSentBertOCAMaxPct}{21.0}
\newcommand{\tevetCxAnVsSentBertOCAMinPct}{3.7}
\newcommand{\tevetDecTestAnPromptGenRho}{+0.932}
\newcommand{\tevetDecTestAnRespGenRho}{+0.924}

\newcommand{\tevetMcDivNugPromptGenCxAnAUC}{0.867}
\newcommand{\tevetMcDivNugPromptGenDdiscAUC}{0.303}
\newcommand{\tevetMcDivNugSetCount}{3069}
\newcommand{\tevetMcDivPromptGenAUC}{0.921}
\newcommand{\tevetMcDivPromptGenCxAnOCA}{0.846}
\newcommand{\tevetMcDivPromptGenCxAnRho}{+0.729}

\newcommand{\tevetMcDivPromptGenSentBertOCA}{0.897}
\newcommand{\tevetMcDivPromptGenSentBertRho}{+0.796}
\newcommand{\tevetMcDivSetCount}{6002}

\newcommand{\modeCountAnMone}{9.3}
\newcommand{\modeCountAnMten}{77.8}

\newcommand{\modeCountMMax}{10}
\newcommand{\modeCountNDraws}{1000}
\newcommand{\modeCountNResponses}{20}
\newcommand{\modeCountPoolSize}{50}

\newcommand{\permSensGPTMisranked}{2}
\newcommand{\permSensHighPerm}{100}
\newcommand{\permSensLowPerm}{3}
\newcommand{\permSensNumScenarios}{5}
\newcommand{\permSensQwenMisranked}{2}

\newcommand{\scenarioMixedGptDCan}{0.52}
\newcommand{\scenarioMixedQwenDCan}{0.41}
\newcommand{\scenarioMultiIncoherentGptDCan}{0.19}
\newcommand{\scenarioMultiIncoherentQwenDCan}{0.29}
\newcommand{\scenarioMultiModeGptDCan}{0.26}
\newcommand{\scenarioMultiModeQwenDCan}{0.19}
\newcommand{\scenarioOneModeGptDCan}{0.10}
\newcommand{\scenarioOneModeQwenDCan}{0.09}
\newcommand{\scenarioPureNoiseGptDCan}{0.04}
\newcommand{\scenarioPureNoiseQwenDCan}{0.05}

%% file: sections/abstract_workshop.tex
Measuring the diversity of creative outputs is central to evaluating post-training mode collapse, comparing decoding strategies, and quantifying creative behavior in both AI and human writing.
We propose a new approach to measuring diversity using in-context learning, of which the ``Decan'' metric, $D_{Ca_n} = C \times a_n$, is the working instance we evaluate: a per-byte score read off the per-token log-probabilities of a base model $\theta$ in a \emph{single forward pass} per permutation, with no embedding model, no reference corpus, and no human labels.
This approach is grounded in information theory, makes use of language model in-context learning to detect a wide range of similarities between any number of inputs, and obviates the need to train a special-purpose model.
The same pipeline scores AI samples and human-written response sets, with diversity treated as a property of (responses, prompt, scoring model).
On Tevet and Berant's human-grounded McDiv benchmark, $D_{Ca_n}$ reaches OCA \tevetMcDivPromptGenCxAnOCA{} on the McDiv prompt\_gen set where it performs best, behind the strongest neural baseline reported in Tevet and Berant (SentBERT, \tevetMcDivPromptGenSentBertOCA{}).
On the OLMo-2-7B post-training pipeline, $D_{Ca_n}$ drops monotonically across the base $\to$ SFT $\to$ DPO $\to$ RLVR stages, detecting the type of diversity loss that creative-writing applications care about.

%% file: sections/01_motivation_workshop.tex
\section{Motivation}\label{sec:motivation}

\begin{figure*}[tbp]
\centering
\begin{tikzpicture}[
    font=\footnotesize,
    box/.style={draw, rounded corners=2pt, align=center, inner sep=3pt},
    stage/.style={box, fill=gray!8},
    cond/.style={box, fill=orange!10},
    uncond/.style={box, fill=blue!8},
    final/.style={box, fill=green!10, very thick},
    arrow/.style={-{Latex[length=1.8mm]}, thick},
    node distance=2mm and 4mm,
]

% --- INPUT + FORMAT (combined, top) ---
\node[stage, text width=0.92\textwidth] (input) {
    \textbf{Input:} prompt $p$, responses $r_1,\ldots,r_n$ from policy $\pi$.\\[2pt]
    \textbf{Format:} \texttt{p\textbackslash n\textbackslash nResponse A: $r_{\sigma(1)}$\textbackslash n\textbackslash nResponse B: $r_{\sigma(2)}$\,\ldots}\\[2pt]
    \textbf{Tokenize:} $[\texttt{Resp}][\texttt{onse}][\texttt{ A}][\texttt{:}][\texttt{ Rain}][\texttt{ falls}]\cdots$
};

% --- CONDITIONAL TRACK (LEFT) ---
\node[cond, below left=5mm and 1mm of input.south, text width=0.45\textwidth, anchor=north east] (cond) {
    \textbf{Conditional track} \;{\scriptsize ($\times\,n_\sigma$ permutations)}\\[2pt]
    Feed concatenated context to base model $\theta$ in one forward pass.\\
    For each response slot $k$:
    \[ a_k^\sigma \;=\; -\!\!\sum_{t \in r_{\sigma(k)}}\!\!\log_2 \theta(t \mid \mathrm{prev}) \]
    Divide by $\|r_{\sigma(k)}\|$ \emph{(per-byte)}.\\
    Average over $\sigma$ to get $\bar{a}_k$ curve.\\[2pt]
    Last point: $a_n \approx a_\infty$
};

% --- UNCONDITIONAL TRACK (RIGHT) ---
\node[uncond, below right=5mm and 1mm of input.south, text width=0.45\textwidth, anchor=north west] (unc) {
    \textbf{Unconditional track} \;{\scriptsize ($n$ separate passes)}\\[2pt]
    Feed each $r_i$ alone (with prompt $p$) to $\theta$.
    \[ h(r_i\mid p) \;=\; -\tfrac{1}{\|r_i\|}\log_2 \theta(r_i \mid p) \]
    \emph{(per-byte)} Take the geometric mean:
    \[ C \;=\; 2^{-\,\frac{1}{n}\sum_i h(r_i\mid p)} \]
};

% Arrows from input down to each track
\draw[arrow] (input.south) -- ++(0,-1mm) -| (cond.north);
\draw[arrow] (input.south) -- ++(0,-1mm) -| (unc.north);

% --- FINAL FORMULA ---
\node[final, below=4mm of input, text width=0.62\textwidth, yshift=-49mm] (out) {
    \textbf{Diversity score:}\quad $D_{Ca_n} \;=\; C \times a_n$\quad{\scriptsize(bits/byte)}
};

% Arrows from each track straight down into the final box
\draw[arrow] (cond.south) -- (cond.south |- out.north);
\draw[arrow] (unc.south)  -- (unc.south |- out.north);

\end{tikzpicture}
\caption{The diversity-metric pipeline.
Given prompt $p$ and $n$ responses from policy $\pi$, we format them with response labels (``Response A:'', ``Response B:'', \ldots) and tokenize.
The \textcolor{orange!70!black}{\textbf{conditional track}} (left) feeds the concatenated context to the base model $\theta$ in a single forward pass per permutation $\sigma$, extracts per-response total surprise, divides by each response's UTF-8 byte count, and averages over permutations to get $\bar a_k$.
Its last point is $a_n$.
The \textcolor{blue!70!black}{\textbf{unconditional track}} (right) scores each $r_i$ independently against the prompt alone, again per-byte; the geometric mean of these surprises defines coherence $C = 1/\mathrm{PPL}_\theta(\pi, p)$, the reciprocal of the geometric-mean per-byte perplexity.
The product $D_{Ca_n} = C \times a_n$ measures plausibility-weighted residual diversity (bits/byte).
Formal definitions: Sections~\ref{sec:setup-notation}, \ref{sec:progressive-conditioning}, \ref{sec:coherence}, \ref{sec:scalar}.}
\label{fig:pipeline}
\end{figure*}
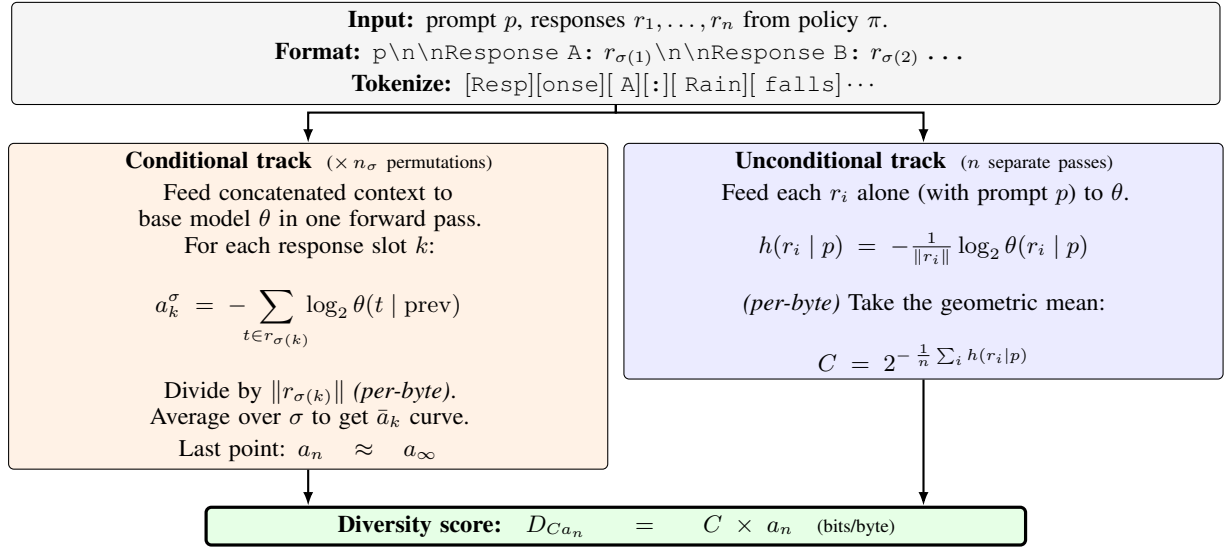

The possibility for diverse outputs is necessary, but not sufficient, for creativity.
Machine learning researchers that study creativity
%Creativity researchers
routinely need to compare the diversity of outputs across generation processes: post-training stages of the same base model (mode collapse), decoding strategies (what does temperature trade off against?), prompting interventions (which prompts have a wide response distribution?), and human-AI hybrid pipelines (would AI use cause human-AI writing to be less creative?).
Existing diversity metrics rely on embedding distances \citep{du2019boosting} or surface-level $n$-gram statistics, both of which lack a human-like ability to recognize arbitrary patterns when those patterns differ from their training data or surface similarities, respectively (\ifworkshop Section\else Appendix\fi~\ref{sec:related}).
A complementary line of work applies information theory directly: Maximum Mutual Information decoding \citep{li2016diversity} and Adversarial Information Maximization \citep{zhang2018aim} optimise pairwise mutual information between input and response, but as a training or decoding objective rather than an evaluation-time diagnostic.
We propose a new approach to measuring diversity using in-context learning, of which the ``Decan'' metric $D_{Ca_n} = C \times a_n$ is the working instance we evaluate: it scores AI samples and human-written response sets through the same pipeline of per-byte log-probabilities of a base model $\theta$, in a \emph{single forward pass} over the concatenated responses (Figure~\ref{fig:pipeline}).
There is no embedding model, no reference corpus, no human labels, no auxiliary classifier.
The approach uses only the per-token probabilities $\theta$ already produces.

The intuition is that if responses from a policy $\pi$ are diverse, seeing one should not help $\theta$ predict the next; if they are repetitive or constrained to a few modes, conditioning on earlier responses should sharply reduce $\theta$'s surprise at later ones.
We exploit the in-context learning capability described by \citet{brown2020language}, applied here as a measurement lens rather than as a few-shot-prompting setup.
The progressive conditional surprise curve $a_k$ (the per-byte cross-entropy of the $k$-th response given the previous $k{-}1$) captures this signal directly, and its last point is $a_n$ (Section~\ref{sec:progressive-conditioning}).
A separate coherence term $C = 1/\mathrm{PPL}_\theta(\pi, p)$, the reciprocal of the geometric-mean per-byte perplexity that $\theta$ assigns to each response individually, prevents pure noise from registering as ``diverse'' (Section~\ref{sec:coherence}).
The product $D_{Ca_n} = C \times a_n$ is the working scalar we adopt; it is plausibility-weighted residual diversity in bits per byte.
We validate it against Tevet and Berant's human-grounded diversity benchmark (Section~\ref{sec:tevet}) and on the OLMo-2-7B post-training pipeline (Section~\ref{sec:rlhf-experiment}), and release a working open-source implementation alongside.\footnote{Code and data: \projectGithubUrl}
The metric measures diversity as $\theta$ perceives it: outputs differing only in ways $\theta$ cannot distinguish in context appear less diverse.
This relativity gives the metric the potential to tighten as base models improve (see Appendix~\ref{app:qwen3-comparison} for a preliminary scaling experiment).

%% file: sections/02_setup.tex
\section{Setup and Notation}\label{sec:setup-notation}

We use the following notation:
\begin{itemize}[nosep]
    \item $p$ be a prompt,
    \item $\pi$ be the policy under evaluation,
    \item $r_1, r_2, \ldots, r_n \sim \pi(\cdot \mid p)$ be $n$ i.i.d.\ responses sampled from $\pi$,
    \item $\theta$ be a trusted base model from which we can obtain per-token log-probabilities,
    \item $|r|_{\mathrm{tok}}$ denote the number of tokens in response $r$,
    \item $\|r\|$ denote the number of bytes in the UTF-8 encoding of response $r$.
\end{itemize}

We define the \textbf{cross-entropy} (total surprise) of a response $r$ under $\theta$:
\begin{equation}\label{eq:total-xent}
    -\log_2 \theta(r \mid p) = \sum_{t=1}^{|r|_{\mathrm{tok}}} -\log_2 \theta(r^t \mid r^{<t}, p)
\end{equation}
where $r^t$ is the $t$-th token and $r^{<t}$ the preceding tokens.
Units: bits.\footnote{Throughout, ``bits'' refers to self-information in $\log_2$ units, identical to the ``shannon'' (Sh) of IEC 80000-13.
We use ``bits'' to match standard usage in the information-theory literature.} This is the total surprise of the response under $\theta$, a function of the string and of $\theta$'s distribution but not of $\theta$'s tokenizer (since the chain rule yields the same total regardless of how the sequence is factored).

Similarly, the \textbf{conditional cross-entropy} given previous responses $r_{<k} = (r_1, \ldots, r_{k-1})$ is $-\log_2 \theta(r_k \mid r_{<k}, p)$ (bits).

For the coherence term and diversity scores (Sections~\ref{sec:coherence} and \ref{sec:scalar}), we also use the \textbf{per-byte cross-entropy rate}:
\begin{equation}\label{eq:pertok}
    h_\theta(r \mid p) = \frac{-\log_2 \theta(r \mid p)}{\|r\|}
\end{equation}
Units: bits/byte.
We adopt this per-byte rate because it works better than total bits in our experiments; we have not investigated why.
Normalising by byte count rather than token count keeps the rate independent of $\theta$'s tokenizer when comparing base models with different vocabularies.

In practice, computing $\theta(r_k \mid r_{<k}, p)$ requires feeding $\theta$ a formatted context containing the prompt and all previous responses (see Section~\ref{sec:practical}).

%% file: sections/03_method_workshop.tex
\section{Method}\label{sec:method}

This section defines the three quantities used in the main body: the progressive conditional surprise curve $a_k$, the coherence term $C$, and the scalar score $D_{Ca_n} = C \times a_n$.
The full information-theoretic motivation, the alternative excess-entropy summary $E$, and the cross-mode learning analysis appear in Appendices~\ref{app:excess-entropy}--\ref{app:qwen3-comparison}.

\subsection{Progressive Conditional Surprise}\label{sec:progressive-conditioning}

Given prompt $p$ and $n$ responses $r_1, \ldots, r_n$ from policy $\pi$, define
\begin{equation}\label{eq:ak}
    a_k = -\log_2 \theta(r_k \mid r_{<k}, p) \quad \text{for } k = 1, \ldots, n,
\end{equation}
the total surprise of the $k$-th response under base model $\theta$ given the previous $k{-}1$ responses.
The sequence $(a_1, \ldots, a_n)$ is the \textbf{progressive conditional surprise curve}.
We normalize each $a_k$ by the byte count $\|r_k\|$ to get a per-byte (bits/byte) curve that is independent of $\theta$'s tokenizer.
All quantities below are per-byte unless otherwise stated.

The intuition is that as $\theta$ sees more responses from $\pi$, the conditional surprise drops if the responses share patterns ($\theta$'s in-context learning picks up on modes, stylistic regularities, topic patterns) and stays flat if they do not.
For a policy with rich coherent diversity the curve declines and levels off at a positive floor; for a policy producing one repeated mode the curve drops sharply to near zero; for pure noise the curve stays roughly constant at a high value.

In practice we average $a_k$ over uniformly random permutations of the response ordering (so each position averages over all responses and the curve reflects only how $\theta$'s predictions improve with more context, not which response happened to land in slot $k$).
We take the metric value to be the last observed point $a_n$.
Equivalently, $a_n = H_\theta(r_n \mid p) - I_\theta(r_n; r_1, \ldots, r_{n-1} \mid p)$: the response's individual surprise under $\theta$ minus how much the prior responses make it predictable.
High $a_n$ therefore requires $r_n$ to be both individually surprising and not made predictable by the others, two properties one would want a diversity score to reward.
No fitting or extrapolation step is involved.
The curve $a_k$ is itself informative: its shape distinguishes a one-mode collapse (sharp drop to a low floor) from richer diversity (gradual decline to a higher floor) in ways the endpoint alone obscures.

\subsection{The Coherence Term}\label{sec:coherence}

The curve alone is fooled by pure noise: $\theta$ can never predict random tokens from one another, so $a_n$ stays at the high unconditional surprise level and noise would dominate the metric.
The distinguishing signal lies in how plausible $\theta$ finds each individual response, independent of the others.

Let $h_\theta(r \mid p) = -\frac{1}{\|r\|}\log_2 \theta(r \mid p)$ be the per-byte cross-entropy of response $r$ under $\theta$ conditioned on the prompt alone.
Define \textbf{coherence} as the geometric mean of the per-byte probabilities:
\begin{equation}\label{eq:coherence}
    C = 2^{-\frac{1}{n}\sum_{i=1}^{n} h_\theta(r_i \mid p)} = \frac{1}{\mathrm{PPL}_\theta(\pi, p)},
\end{equation}
the reciprocal of the geometric-mean per-byte perplexity\footnote{In our Tevet evaluation with Qwen2.5-3B, the central 90\% (5th to 95th percentile) of per-response bits-per-byte values across \tevetCoherenceRespCount{} responses span $[\tevetCoherenceBpbLow, \tevetCoherenceBpbHigh]$, giving $C \in [\tevetCoherenceCLow, \tevetCoherenceCHigh]$; the mean per-set $C$ across \tevetCoherenceSetCount{} response sets is \tevetCoherenceMeanC.} that $\theta$ assigns to the responses individually.\footnote{For scale, the bits/byte score is the average per-byte cross-entropy $\bar{\ell}$, related to coherence by $C = 2^{-\bar{\ell}}$. GPT-3 davinci on The Pile \citep{gao2020pile} scored $\sim$0.72 bits/byte and GPT-2 XL $\sim$1.05 bits/byte, giving $C \approx 0.61$ and $C \approx 0.49$ respectively; stronger base models reach lower bits/byte and correspondingly higher $C$.}
Perplexity is a standard metric of incoherence, and the geometric form is intended to suppress sets containing incoherent responses: a single sample with high per-byte cross-entropy drives $C$ toward zero, limiting the rescue effect of any single fluent response on an otherwise incoherent set.

\subsection{The Diversity Score}\label{sec:scalar}

The scalar score is the product:
\begin{equation}\label{eq:D-Can}
    D_{Ca_n} = C \times a_n \quad \text{(bits/byte).}
\end{equation}
It can be read as ``how many bits of surprise per byte remain after $\theta$ has learned what it can from the other responses, weighted by output plausibility.''
The intended edge-case behaviors are summarized in Table~\ref{tab:edge-cases}, alongside empirical $D_{Ca_n}$ on synthetic instances of each scenario under two base models (full per-metric breakdown in Appendix~\ref{sec:scenario-validation}, Table~\ref{tab:scenarios}).

\begin{table*}[t]
\centering
\small
\caption{Intended edge-case behavior of $D_{Ca_n} = C \times a_n$ on the five synthetic scenarios (Appendix~\ref{sec:scenario-validation}), with empirical $D_{Ca_n}$ values under GPT-2 (124M) and Qwen2.5-3B base.
The product correctly suppresses pure noise and incoherent multi-mode (via $C$) and one-mode sets (via $a_n$); multi-mode coherent is the intended winner.
Empirical numbers show smaller $a_n$ for multi-mode coherent than this story predicts: with only 3 distinct modes in a set of 10 responses, $\theta$'s in-context learning identifies them and $a_n$ is no longer high. In such ambiguous cases reporting the full $a_k$ curve is more principled than the scalar $a_n$: a different choice of $n$ would reveal the diversity, and showing the curve sidesteps having to pick that $n$ in advance. Appendix~\ref{sec:mode-count-scaling} shows $a_n$ rising with mode count.
Mixed empirically scores highest (via a high $a_n$) rather than the predicted ``mid'' position; reweighting strategies are discussed in Section~\ref{sec:exp-discussion}. On Qwen2.5-3B, multi-mode incoherent also outranks multi-mode coherent; Section~\ref{sec:scalar} discusses.}
\label{tab:edge-cases}
\begin{tabular}{@{}lccccc@{}}
\toprule
                      & \multicolumn{3}{c}{Predicted}            & \multicolumn{2}{c}{Empirical $D_{Ca_n}$} \\
\cmidrule(lr){2-4} \cmidrule(lr){5-6}
Scenario              & $C$  & $a_n$ & $D_{Ca_n}$              & GPT-2  & Qwen2.5-3B \\
\midrule
Pure noise            & low  & high  & low (via $C$)           & \scenarioPureNoiseGptDCan{}        & \scenarioPureNoiseQwenDCan{}        \\
Multi-mode incoherent & low  & high  & low (via $C$)           & \scenarioMultiIncoherentGptDCan{}  & \scenarioMultiIncoherentQwenDCan{}  \\
Multi-mode coherent   & high & high  & \textbf{high}           & \scenarioMultiModeGptDCan{}        & \scenarioMultiModeQwenDCan{}        \\
One-mode              & high & low   & low (via $a_n$)         & \scenarioOneModeGptDCan{}          & \scenarioOneModeQwenDCan{}          \\
Mixed                 & mid  & high  & mid                     & \scenarioMixedGptDCan{}            & \scenarioMixedQwenDCan{}            \\
\bottomrule
\end{tabular}
\end{table*}

The score depends on $p$, $\pi$'s responses, and $\theta$.
Diversity is therefore a property of (responses, prompt, scoring model), not of the policy in isolation: the same response set can score differently under a stronger or weaker $\theta$.
This is by design: $\theta$'s in-context learning capability is the lens through which diversity is measured, giving the score the potential to improve as base models improve (see Appendix~\ref{app:qwen3-comparison} for a preliminary scaling experiment).
For example, Table~\ref{tab:edge-cases} shows Qwen2.5-3B scoring multi-mode incoherent (\scenarioMultiIncoherentQwenDCan{}) above multi-mode coherent (\scenarioMultiModeQwenDCan{}), reversing our predicted ranking; Figure~\ref{fig:scenario-curves} shows Qwen's $\bar{a}_k$ curve dropping over $k$ on the multi-mode incoherent scenario, indicating that its in-context learning recognises the shared template structure as a pattern despite the within-response scrambling.
GPT-2's weaker in-context learning fits the predicted ordering on this row better (\scenarioMultiIncoherentGptDCan{} for incoherent vs.\ \scenarioMultiModeGptDCan{} for coherent), but we still recommend stronger base models when possible -- sophisticated patterns are exactly what an ICL-based metric should be able to detect.
When comparing policies across a prompt suite, $D_{Ca_n}$ can be averaged over prompts or differenced; see Appendix~\ref{app:aggregation}.

%% file: sections/10_related.tex
\section{Related Work}\label{sec:related}

\paragraph{Sampling diversity metrics.} Work on decoding-time diversity typically operates at the surface level: $n$-gram overlap \citep{li2016diversity} and self-BLEU \citep{zhu2018texygen}.
Our approach operates at the distributional level and can distinguish policies that produce lexically varied but semantically redundant outputs.

\paragraph{Diversity evaluation benchmarks.}
\citet{tevet2020evaluating} introduced a systematic framework for evaluating diversity metrics, with their McDiv benchmark providing human-labeled response sets at two diversity levels.
We use McDiv for validation (Section~\ref{sec:tevet}), though we identify a construction confound in how its low-diversity sets are produced (Appendix~\ref{app:confound}).
More recent work has revisited the meta-evaluation problem.
NoveltyBench \citep{zhang2025noveltybench} defines diversity through pairwise functional equivalence rather than binary labels, training a DeBERTa classifier to group generations into equivalence classes and computing a \emph{Distinct}$_k$ score (the number of meaningfully different outputs in $k$ samples), conceptually close to our mode count.
However, their ground truth is itself a trained classifier (79\% accuracy), making it unclear whether correlation with their labels validates a metric or merely measures agreement between two model-based scores.
\citet{zhang2025commonsense} conduct a meta-evaluation of diversity metrics for constrained commonsense generation, using GPT-4o as an annotator in place of crowd workers.
\citet{guo2024linguistic} benchmark linguistic diversity of LLMs, building on Tevet and Berant's framework with a broader set of generation tasks.

\paragraph{Perplexity.} The coherence term $C = 1/\mathrm{PPL}$ (Section~\ref{sec:coherence}) connects our framework to the standard language modeling evaluation metric.
The primary diversity score $D_{Ca_n} = C \times a_n$ operates in bits/byte, weighting the residual surprise floor by output plausibility.

\paragraph{Coherence as a signal.} Our coherence term $C$ uses the base model's predictive distribution as an unsupervised quality signal.
This connects to a broader line of work using model-internal coherence without external supervision: \citet{qiu2026selfimprovement} show theoretically that feedback-free self-improvement methods work by optimizing coherence (compressibility of context-to-behavior mappings), and \citet{wen2025unsupervised} use internal coherence maximization to elicit capabilities from language models.
Our framework leverages a related insight: the base model's ability to compress responses in context provides a meaningful signal about their diversity structure.

\paragraph{Embedding-based diversity.} \citet{du2019boosting} introduced the embedding-based approach, clustering sentence embeddings of generated responses with $k$-means and reporting cluster inertia as the diversity score; subsequent work has explored alternative geometric statistics over embedded responses.
These approaches are complementary to ours: they capture semantic distance in a learned representation space, while our metric captures statistical independence under a generative model.
The approaches may disagree when $\theta$'s in-context learning captures structure invisible to the embedding model, or vice versa.

\paragraph{Output collapse in RL training.} \citet{wang2025ragen} document an ``Echo Trap'' pattern during multi-turn agent RL: early-stage agents produce varied symbolic reasoning, then collapse to deterministic templates as training proceeds.
They detect this collapse using within-prompt reward standard deviation as an early-warning signal, a proxy that relies on the reward distinguishing degenerate outputs from diverse ones.
The $a_k$ framework provides a text-level alternative: collapse should appear as a drop in $a_n$ on the agent's own outputs, independent of reward structure. A downside of using the $a_k$ framework is that it is unconnected from a downstream task. Responses can both be "diverse" as measured by $a_k$-based metrics and fail uniformly.

\paragraph{Excess entropy.} The $a_k$ curve also admits an excess-entropy summary related to the computational mechanics literature \citep{crutchfield2003regularities}; see Appendix~\ref{app:excess-entropy}.
We found it empirically inferior to $D_{Ca_n}$ on Tevet and Berant's diversity-eval benchmark \citep{tevet2020evaluating}.

%% file: sections/07_5_tevet_workshop.tex
\section{Tevet--Berant Diversity-Eval: Human-Grounded Validation}\label{sec:tevet}

A diversity metric should be tested against human judgements on a standardized eval; \citet{tevet2020evaluating} provide both.
Their McDiv and ConTest datasets pair human-grounded diversity labels with a fixed comparison protocol (Spearman $\rho$ and OCA: \emph{optimal classification accuracy}, the best accuracy achievable by a one-dimensional threshold separating low- and high-diversity sets), the standard methodology for ranking diversity metrics against human judgements.
The response sets are written by Mechanical Turk workers, and for both McDiv\_nuggets and ConTest the high/low label is fixed by the construction protocol itself: a worker writes five different continuations (the high-diversity set), then self-selects one of those continuations and paraphrases it five times preserving content but varying form (the low-diversity set; see Appendix~\ref{app:confound} for the full protocol and a construction-confound caveat).
We run $D_{Ca_n}$ through this evaluation pipeline.

\paragraph{Setup.} We use Qwen2.5-3B (base) with 50 permutations in completion format (Appendix~\ref{sec:formatting}) on the released splits: \tevetMcDivSetCount{} McDiv sets (no\_hds), \tevetMcDivNugSetCount{} McDiv\_nuggets sets (no\_hds), and \tevetConTestSetCount{} ConTest sets (with\_hds), each with 5 worker-written responses per set and a binary high/low diversity label.
We follow Tevet and Berant in reporting Spearman $\rho$ and OCA, and additionally report ROC AUC (the natural metric for binary classification).

\begin{table*}[t]
\centering
\caption{ConTest results: Spearman $\rho$ and OCA between a set's diversity class and each metric score.
Baselines from Tevet's pre-computed metrics; $C \!\times\! a_n$ and $a_n$ are ours (Qwen2.5-3B, 50 permutations, per-byte).
Best result per task in bold.}
\label{tab:contest}
\small
\input{results/tables/contest_rho_oca.tex}
\end{table*}

\paragraph{Headline result.} On the binary tasks $D_{Ca_n}$ clearly tracks diversity, with OCA between \tevetCxAnVsSentBertOCAMinPct\% and \tevetCxAnVsSentBertOCAMaxPct\% behind SentBERT (Table~\ref{tab:contest}).
On McDiv prompt\_gen, where $D_{Ca_n}$ performs best, it reaches $\rho = \tevetMcDivPromptGenCxAnRho{}$ and OCA = \tevetMcDivPromptGenCxAnOCA{}, against SentBERT's $\rho = \tevetMcDivPromptGenSentBertRho{}$ and OCA = \tevetMcDivPromptGenSentBertOCA{}; the corresponding ROC AUC is \tevetMcDivPromptGenAUC{}.
The headline takeaway is that an ICL-based diversity metric reaches close to a sentence-embedding baseline on a human-grounded eval, with no embedding model, no reference corpus, and no human labels in the metric itself.

\paragraph{DecTest, included for completeness.} Tevet and Berant also release DecTest (Table~\ref{tab:dectest}), a diagnostic where the ``label'' is the sampling temperature used to generate each response set rather than a human judgement.
We include it for parity with the released benchmark only: we do not believe that temperature labels map onto creative/meaningful output diversity, consistent with \citet{peeperkorn2024temperature}'s finding that the relationship between sampling temperature and creativity is more nuanced and weaker than the ``creativity parameter'' framing suggests.
The raw $a_n$ achieves $\rho = \tevetDecTestAnPromptGenRho{}$ on prompt\_gen and $\rho = \tevetDecTestAnRespGenRho{}$ on resp\_gen, matching or exceeding all baselines.

\begin{table}[t]
\centering
\caption{DecTest results: Spearman $\rho$ between sampling temperature and each metric score (1000 samples, no\_hds).
Reported for parity with Tevet and Berant; the labels here are sampling temperatures rather than human judgements, so DecTest is not part of the framing for $D_{Ca_n}$.
That $a_n$ tracks temperature is mechanically unsurprising: $a_n$ is a conditional-entropy quantity, and sampling temperature directly scales the entropy of the policy, so a positive $\rho$ here is a sanity check rather than evidence about creative diversity; \citet{peeperkorn2024temperature} likewise find only a weak relationship between temperature and creativity in LLM outputs.}
\label{tab:dectest}
\small
\input{results/tables/dectest_rho.tex}
\end{table}

\paragraph{Caveats.} The McDiv\_nuggets construction protocol introduces a confound: low-diversity sets are paraphrases of specific, dramatic endings that are intrinsically more surprising to the base model than typical high-diversity continuations.
Appendix~\ref{app:confound} documents the mechanism, the per-byte $a_1$ gap, and an evidence-by-length-bin breakdown showing the effect is content-driven not length-driven.
The $C$ weighting in $D_{Ca_n}$ helps on this benchmark partly for that confound-related reason; on settings without such a confound we expect $a_n$ to carry most of the signal.

%% file: results/tables/contest_rho_oca.tex
% Generated by: scripts/analyze_c_ainf.py --run-tag qwen25_completion_v3
% Baseline source: diversity-eval/data/with_metrics/
\begin{tabular}{@{}l rr rr rr@{}}
\toprule
& \multicolumn{2}{c}{prompt\_gen} & \multicolumn{2}{c}{resp\_gen} & \multicolumn{2}{c}{story\_gen} \\
\cmidrule(lr){2-3} \cmidrule(lr){4-5} \cmidrule(lr){6-7}
Metric & $\rho$ & OCA & $\rho$ & OCA & $\rho$ & OCA \\
\midrule
\multicolumn{7}{@{}l}{\textit{ConTest (200, with\_hds)}} \\[2pt]
$C \!\times\! a_n$ (ours) & +0.584 & 0.785 & +0.391 & 0.668 & +0.686 & 0.828 \\
$a_n$ (ours) & +0.444 & 0.715 & +0.274 & 0.641 & +0.387 & 0.684 \\
$C$ (ours) & +0.214 & 0.625 & $-0.001$ & 0.555 & +0.247 & 0.632 \\
SentBERT & \textbf{+0.682} & 0.815 & \textbf{+0.591} & \textbf{0.791} & \textbf{+0.770} & \textbf{0.896} \\
BERTsts & +0.646 & \textbf{0.820} & +0.463 & 0.714 & +0.601 & 0.780 \\
distinct-$n$ & +0.333 & 0.675 & +0.346 & 0.677 & +0.573 & 0.772 \\
\midrule
\multicolumn{7}{@{}l}{\textit{McDiv\_nuggets ($\sim$1K, no\_hds)}} \\[2pt]
$C \!\times\! a_n$ (ours) & +0.636 & 0.785 & +0.345 & 0.649 & +0.317 & 0.634 \\
$a_n$ (ours) & +0.487 & 0.705 & +0.225 & 0.619 & +0.124 & 0.557 \\
$C$ (ours) & +0.138 & 0.567 & +0.082 & 0.545 & +0.251 & 0.643 \\
SentBERT & \textbf{+0.728} & \textbf{0.850} & \textbf{+0.532} & \textbf{0.758} & \textbf{+0.633} & \textbf{0.803} \\
BERTsts & +0.683 & 0.830 & +0.393 & 0.676 & +0.344 & 0.638 \\
distinct-$n$ & $-0.003$ & 0.514 & $-0.002$ & 0.507 & $-0.002$ & 0.510 \\
\midrule
\multicolumn{7}{@{}l}{\textit{McDiv (full, no\_hds, $\sim$2K)}} \\[2pt]
$C \!\times\! a_n$ (ours) & +0.729 & 0.846 & +0.500 & 0.724 & +0.523 & 0.717 \\
$a_n$ (ours) & +0.617 & 0.781 & +0.432 & 0.698 & +0.402 & 0.668 \\
$C$ (ours) & +0.138 & 0.565 & $-0.007$ & 0.512 & +0.171 & 0.594 \\
SentBERT & \textbf{+0.796} & \textbf{0.897} & \textbf{+0.678} & \textbf{0.830} & \textbf{+0.753} & \textbf{0.867} \\
BERTsts & +0.780 & 0.893 & +0.614 & 0.781 & +0.571 & 0.740 \\
distinct-$n$ & +0.476 & 0.746 & +0.517 & 0.738 & +0.535 & 0.744 \\
\bottomrule
\end{tabular}

%% file: results/tables/dectest_rho.tex
% Generated by: scripts/analyze_c_ainf.py --run-tag qwen25_completion_v3
% Baseline source: diversity-eval/data/with_metrics/
\begin{tabular}{@{}l rrr@{}}
\toprule
Metric & prompt\_gen & resp\_gen & story\_gen \\
\midrule
$C \!\times\! a_n$ (ours) & +0.847 & +0.771 & +0.763 \\
$a_n$ (ours) & \textbf{+0.932} & \textbf{+0.924} & \textbf{+0.779} \\
$C$ (ours) & $-0.727$ & $-0.813$ & $-0.547$ \\
distinct-$n$ & +0.917 & +0.894 & +0.758 \\
BERTScore & +0.878 & +0.874 & +0.694 \\
SentBERT & +0.747 & +0.801 & +0.645 \\
cos-sim & +0.873 & +0.895 & +0.712 \\
\bottomrule
\end{tabular}

%% file: sections/07_6_rlhf_workshop.tex
% ============================================================================
% Workshop variant of the OLMo-2-7B RLHF case study.
% Differences from sections/07_6_rlhf.tex:
%   - Promoted from \subsection to \section (no §7 Experiments parent in
%     the workshop wrapper).
%   - One-paragraph framing intro added so this section is read as the
%     AI-side anchor (parallel to Tevet as the Human-side anchor).
%   - Cross-model comparison trimmed to a one-sentence pointer to the
%     appendix; full table and discussion live in sections/07_6_rlhf.tex
%     which is appendix-only in the workshop wrapper.
% ============================================================================

% Load generated macros BEFORE the section body so prose references resolve.
\IfFileExists{results/rlhf_experiment/paper_macros_rlhf.tex}{%
  \input{results/rlhf_experiment/paper_macros_rlhf.tex}}{}
\IfFileExists{results/rlhf_experiment/paper_macros_rlhf_lenmatched.tex}{%
  \input{results/rlhf_experiment/paper_macros_rlhf_lenmatched.tex}}{}
\IfFileExists{results/rlhf_experiment/tables/cross_model_macros.tex}{%
  \input{results/rlhf_experiment/tables/cross_model_macros.tex}}{}

\section{OLMo-2-7B Post-Training: AI-Side Validation}\label{sec:rlhf-experiment}

The Tevet evaluation in Section~\ref{sec:tevet} validates $D_{Ca_n}$ against human-grounded labels.
This section evaluates $D_{Ca_n}$ on real policies: we sample from four post-training stages of the same base model and ask whether the metric detects the diversity loss widely attributed to RLHF \citep{kirk2023rlhf,zhang2025verbalizedsampling,padmakumar2023writingreduces}.
The same scoring pipeline applies to both human-written response sets (Section~\ref{sec:tevet}) and the policy-sampled sets here, so $\theta$ is held fixed across the two evaluations.

\paragraph{Setup.} We sample from the four released stages of the OLMo-2-1124-7B pipeline \citep{olmo2024olmo2}: the pretrained base model, its SFT checkpoint, the DPO checkpoint trained on preference data, and the final RLVR-tuned Instruct checkpoint.
On two prompt sets, \olmoAlpacaN{} AlpacaFarm evaluation prompts \citep{dubois2024alpacafarm} (seed=42 subsample of the 805-prompt set later adopted by AlpacaEval) and \olmoNbCuratedN{} curated NoveltyBench prompts \citep{zhang2025noveltybench}, we draw $K{=}10$ responses per prompt per stage (temperature 1.0, top-$p$ 1.0, $\texttt{max\_new\_tokens}{=}100$).
Base runs raw; SFT / DPO / Instruct receive the prompt through their own chat template.
Because per-byte cross-entropy in a causal LM decreases with within-response context, we truncate every (stage, prompt) tuple's responses to a common per-prompt UTF-8 byte length before scoring,\footnote{Truncation is applied to the response \emph{string} via \texttt{text.encode("utf-8")[:N].decode("utf-8", errors="ignore")} and the truncated string is then re-tokenised for the forward pass. We deliberately do not truncate at token boundaries: $a_n$ is per-byte, so we need a fixed byte denominator across stages, but tokens-per-byte is tokenizer-dependent and varies across responses with the same byte budget. That variation is harmless because both the bits-numerator and the byte-denominator are computed on the same truncated string.} and drop prompts where that common length falls below 50 bytes (\olmoLmAlpacaN{}/\olmoAlpacaN{} AlpacaEval and \olmoLmNbCuratedN{}/\olmoNbCuratedN{} NB-curated prompts retained; see Section~\ref{sec:limitations}).
We score each (stage, prompt) group with $D_{Ca_n}$ using Qwen2.5-3B as $\theta$ and 25 permutations.
For comparison we also compute Kirk et al.'s EAD and distinct-$n$ (averaged over $n{=}1\ldots5$) and a SentBERT-embedding \citep{reimers2019sbert} similarity-to-diversity reduction.

\paragraph{Hypotheses.} We pre-register three one-sided paired Wilcoxon signed-rank tests \citep{dror2018hitchhiker} with family-wise Bonferroni correction across the three contrasts \citep{dror2017replicability} ($\alpha = 0.05/3$):

\begin{itemize}[nosep]
    \item \textbf{H1a} $D_{\mathrm{base}} > D_{\mathrm{SFT}}$: SFT narrows toward the helpful-assistant style.
    \item \textbf{H1b} $D_{\mathrm{SFT}} > D_{\mathrm{DPO}}$: DPO is trained on preference data and inherits its typicality bias \citep{zhang2025verbalizedsampling}.
    \item \textbf{H1c} $D_{\mathrm{base}} > D_{\mathrm{RLVR}}$: cumulative effect of the full post-training pipeline.
\end{itemize}

The DPO-vs-RLVR comparison is reported as an exploratory two-sided contrast (H1$'$): we have no directional prediction for which post-training stage loses more diversity.

\paragraph{Results.}

\begin{table}[h]
\centering
\caption{Per-prompt $D_{Ca_n}$ by OLMo-2-7B stage on the length-matched subset, with pre-registered paired tests. Reported $p$-values for H1a--H1c are Bonferroni-corrected ($\times 3$); the H1$'$ row (marked $^{*}$) is an uncorrected, two-sided exploratory comparison whose direction we had no prior expectation for, and is not a load-bearing claim.}
\label{tab:rlhf-diversity}
\small
\IfFileExists{results/rlhf_experiment/tables_length_matched/rlhf_diversity.tex}{%
  \input{results/rlhf_experiment/tables_length_matched/rlhf_diversity.tex}}{%
  \texttt{[table will appear here after \texttt{5\_analyze\_and\_figures.py} runs]}%
}
\end{table}

Table~\ref{tab:rlhf-diversity} reports per-stage means and the pre-registered Wilcoxon contrasts on both prompt sets.
On AlpacaEval and NoveltyBench-curated alike, $D_{Ca_n}$ decreases monotonically across base $\to$ SFT $\to$ DPO $\to$ Instruct (RLVR), and all three pre-registered contrasts (H1a--H1c) are significant after Bonferroni correction.
\ifworkshop
The underlying $\bar{a}_k$ curves and per-prompt $D_{Ca_n}$ distributions are visualised in Figure~\ref{fig:rlhf-ak-violin}: each later stage's curve lies below the base curve at every $k \geq 2$, and the per-prompt distribution shifts toward lower $D_{Ca_n}$ as the pipeline advances.

\begin{figure*}[h]
\centering
\begin{subfigure}[t]{0.48\textwidth}
    \includegraphics[width=\textwidth]{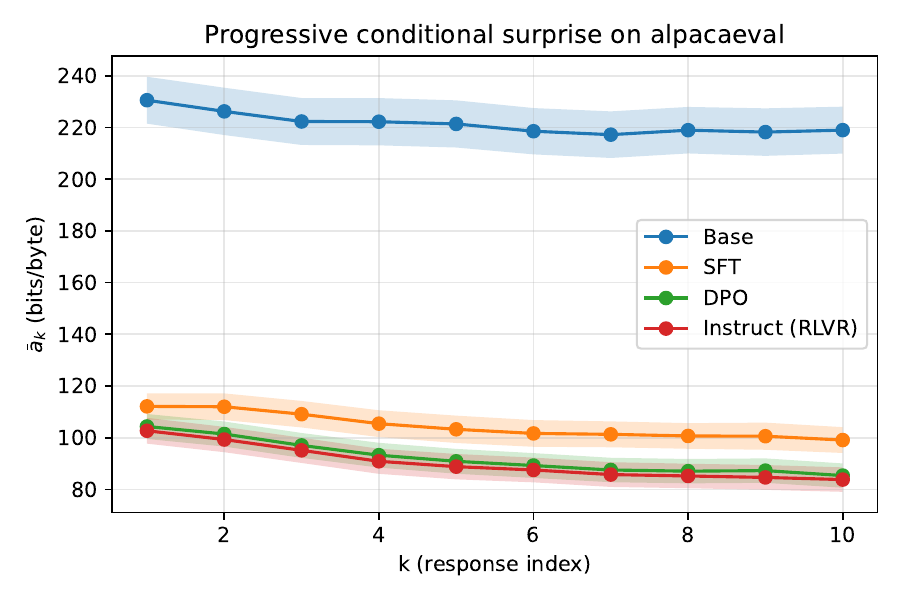}
    \caption{AlpacaEval $\bar{a}_k$ curves (length-matched).}
\end{subfigure}\hfill
\begin{subfigure}[t]{0.48\textwidth}
    \includegraphics[width=\textwidth]{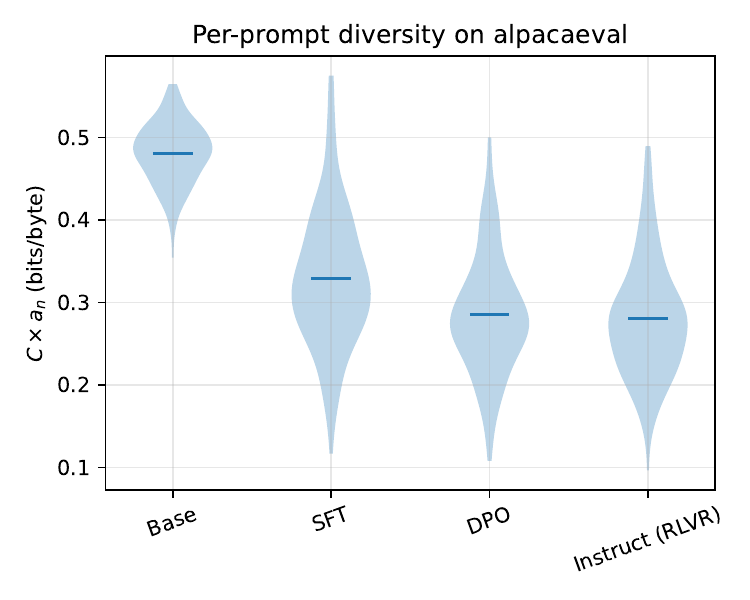}
    \caption{Per-prompt $D_{Ca_n}$ by stage (length-matched).}
\end{subfigure}
\caption{Progressive conditional surprise curves and per-prompt diversity distributions across the four OLMo-2-7B stages on the length-matched AlpacaEval subset. Each later stage's $\bar{a}_k$ curve lies below the base curve at every $k \geq 2$; the per-prompt $D_{Ca_n}$ distribution shifts toward lower values as the pipeline advances.}
\label{fig:rlhf-ak-violin}
\end{figure*}
\else
The underlying $\bar{a}_k$ curves and per-prompt $D_{Ca_n}$ distributions are visualised in Appendix~\ref{app:rlhf-cross-metric} (Figure~\ref{fig:rlhf-ak-violin}): each later stage's curve lies below the base curve at every $k \geq 2$, and the per-prompt distribution shifts toward lower $D_{Ca_n}$ as the pipeline advances.
\fi

\paragraph{Discussion.} The monotone drop across all three pre-registered contrasts is consistent with the post-training-reduces-diversity findings of \citet{kirk2023rlhf,padmakumar2023writingreduces}, now measured by an ICL-based diversity metric that needs neither an embedding model nor a sentence similarity function.
Per-prompt $D_{Ca_n}$ also agrees with the lexical (EAD) and semantic (SentBERT) baselines on the same length-matched subset, confirming the metrics see the same diversity-loss signal (Appendix~\ref{app:rlhf-cross-metric}, Figure~\ref{fig:rlhf-metric-scatter}).
% Cross-model data lives in results/rlhf_experiment/tables/cross_model_macros.tex
% (already \input'd at the top of this file). Within-pipeline OLMo stages:
% \crossModelOlmoBase / \crossModelOlmoSft / \crossModelOlmoDpo / \crossModelOlmoInstruct
% (range 0.485 -> 0.283, drop ~0.202 bits/byte). Frontier instruct cluster on the
% same NoveltyBench-curated prompts: \crossModelQwenFourB, \crossModelQwenEightB,
% \crossModelQwenTwoThreeFiveB, \crossModelLlamaThreeThreeSeventyB,
% \crossModelGptFiveNano, \crossModelGptFive, \crossModelClaudeSonnetFourFive
% (range 0.169 - 0.262, spread ~0.093). The within-pipeline drop is ~2.2x the
% cluster spread; an earlier draft characterised these as "comparable in magnitude",
% which overstated the agreement, so the comparison is deferred to a longer venue
% where the full table can be shown rather than verbally summarised.
\paragraph{Released artefacts.} The \olmoAlpacaN$\,+\,$\olmoNbCuratedN{} prompts $\times$ 4 stages $\times$ $K{=}10$ generations are released as a public dataset under \texttt{results/rlhf\_experiment/} in the project repository.
To our knowledge this is the first public release of $K\!\ge\!10$ sampled responses across a paired SFT/DPO/RL pipeline, filling a gap that blocked direct diversity replication of \citet{kirk2023rlhf} (whose $K{=}16$ generations were not publicly released).

% --- macro fallbacks (only fire if the generated files are absent) ---
\providecommand{\olmoAlpacaN}{200}
\providecommand{\olmoNbCuratedN}{100}
\providecommand{\olmoAlpacaBaseDMean}{??}
\providecommand{\olmoAlpacaSftDMean}{??}
\providecommand{\olmoAlpacaDpoDMean}{??}
\providecommand{\olmoAlpacaInstructDMean}{??}
\providecommand{\olmoAlpacaHoneADz}{??}
\providecommand{\olmoAlpacaHoneAPbonf}{??}
\providecommand{\olmoAlpacaHoneBDz}{??}
\providecommand{\olmoAlpacaHoneBPbonf}{??}
\providecommand{\olmoAlpacaHoneCDz}{??}
\providecommand{\olmoAlpacaHoneCPbonf}{??}
\providecommand{\olmoAlpacaHpaDiff}{??}
\providecommand{\olmoAlpacaHpaDz}{??}
\providecommand{\olmoNbCuratedBaseDMean}{??}
\providecommand{\olmoNbCuratedSftDMean}{??}
\providecommand{\olmoNbCuratedDpoDMean}{??}
\providecommand{\olmoNbCuratedInstructDMean}{??}
\providecommand{\olmoNbCuratedHoneADz}{??}
\providecommand{\olmoNbCuratedHoneAPbonf}{??}
\providecommand{\olmoNbCuratedHoneBDz}{??}
\providecommand{\olmoNbCuratedHoneBPbonf}{??}
\providecommand{\olmoNbCuratedHoneCDz}{??}
\providecommand{\olmoNbCuratedHoneCPbonf}{??}

% Length-matched run macros (only the prompt counts are referenced here)
\providecommand{\olmoLmAlpacaN}{??}
\providecommand{\olmoLmNbCuratedN}{??}

% Cross-model comparison table macros
\providecommand{\crossModelOlmoBase}{??}
\providecommand{\crossModelOlmoSft}{??}
\providecommand{\crossModelOlmoDpo}{??}
\providecommand{\crossModelOlmoInstruct}{??}
\providecommand{\crossModelQwenFourB}{??}
\providecommand{\crossModelQwenEightB}{??}
\providecommand{\crossModelQwenTwoThreeFiveB}{??}
\providecommand{\crossModelLlamaThreeThreeSeventyB}{??}
\providecommand{\crossModelGptFiveNano}{??}
\providecommand{\crossModelGptFive}{??}
\providecommand{\crossModelClaudeSonnetFourFive}{??}

%% file: results/rlhf_experiment/paper_macros_rlhf.tex
% Auto-generated by scripts/rlhf_experiment/5_analyze_and_figures.py

\newcommand{\olmoAlpacaBaseDMean}{0.488}
\newcommand{\olmoAlpacaSftDMean}{0.330}
\newcommand{\olmoAlpacaDpoDMean}{0.290}
\newcommand{\olmoAlpacaInstructDMean}{0.285}
\newcommand{\olmoAlpacaHoneAPbonf}{3.7 \times 10^{-31}}
\newcommand{\olmoAlpacaHoneADz}{1.614}
\newcommand{\olmoAlpacaHoneADiff}{0.158}
\newcommand{\olmoAlpacaHoneBPbonf}{1.0 \times 10^{-15}}
\newcommand{\olmoAlpacaHoneBDz}{0.577}
\newcommand{\olmoAlpacaHoneBDiff}{0.040}
\newcommand{\olmoAlpacaHoneCPbonf}{3.3 \times 10^{-34}}
\newcommand{\olmoAlpacaHoneCDz}{2.526}
\newcommand{\olmoAlpacaHoneCDiff}{0.202}
\newcommand{\olmoAlpacaHpaP}{5.3 \times 10^{-5}}
\newcommand{\olmoAlpacaHpaDz}{0.222}
\newcommand{\olmoAlpacaHpaDiff}{0.005}
\newcommand{\olmoAlpacaN}{200}
\newcommand{\olmoNbCuratedBaseDMean}{0.498}
\newcommand{\olmoNbCuratedSftDMean}{0.384}
\newcommand{\olmoNbCuratedDpoDMean}{0.318}
\newcommand{\olmoNbCuratedInstructDMean}{0.315}
\newcommand{\olmoNbCuratedHoneAPbonf}{1.0 \times 10^{-14}}
\newcommand{\olmoNbCuratedHoneADz}{1.232}
\newcommand{\olmoNbCuratedHoneADiff}{0.113}
\newcommand{\olmoNbCuratedHoneBPbonf}{4.7 \times 10^{-13}}
\newcommand{\olmoNbCuratedHoneBDz}{0.897}
\newcommand{\olmoNbCuratedHoneBDiff}{0.066}
\newcommand{\olmoNbCuratedHoneCPbonf}{1.5 \times 10^{-17}}
\newcommand{\olmoNbCuratedHoneCDz}{2.054}
\newcommand{\olmoNbCuratedHoneCDiff}{0.183}
\newcommand{\olmoNbCuratedHpaP}{0.016}
\newcommand{\olmoNbCuratedHpaDz}{0.069}
\newcommand{\olmoNbCuratedHpaDiff}{0.004}
\newcommand{\olmoNbCuratedN}{100}

%% file: results/rlhf_experiment/paper_macros_rlhf_lenmatched.tex
% Auto-generated by scripts/rlhf_experiment/5_analyze_and_figures.py

\newcommand{\olmoLmAlpacaBaseDMean}{0.481}
\newcommand{\olmoLmAlpacaSftDMean}{0.329}
\newcommand{\olmoLmAlpacaDpoDMean}{0.286}
\newcommand{\olmoLmAlpacaInstructDMean}{0.281}
\newcommand{\olmoLmAlpacaHoneAPbonf}{3.4 \times 10^{-24}}
\newcommand{\olmoLmAlpacaHoneADz}{1.615}
\newcommand{\olmoLmAlpacaHoneADiff}{0.151}
\newcommand{\olmoLmAlpacaHoneBPbonf}{1.7 \times 10^{-13}}
\newcommand{\olmoLmAlpacaHoneBDz}{0.675}
\newcommand{\olmoLmAlpacaHoneBDiff}{0.044}
\newcommand{\olmoLmAlpacaHoneCPbonf}{5.2 \times 10^{-26}}
\newcommand{\olmoLmAlpacaHoneCDz}{2.425}
\newcommand{\olmoLmAlpacaHoneCDiff}{0.200}
\newcommand{\olmoLmAlpacaHpaP}{0.004}
\newcommand{\olmoLmAlpacaHpaDz}{0.227}
\newcommand{\olmoLmAlpacaHpaDiff}{0.005}
\newcommand{\olmoLmAlpacaN}{150}
\newcommand{\olmoLmNbCuratedBaseDMean}{0.481}
\newcommand{\olmoLmNbCuratedSftDMean}{0.369}
\newcommand{\olmoLmNbCuratedDpoDMean}{0.312}
\newcommand{\olmoLmNbCuratedInstructDMean}{0.303}
\newcommand{\olmoLmNbCuratedHoneAPbonf}{4.1 \times 10^{-8}}
\newcommand{\olmoLmNbCuratedHoneADz}{1.212}
\newcommand{\olmoLmNbCuratedHoneADiff}{0.112}
\newcommand{\olmoLmNbCuratedHoneBPbonf}{2.8 \times 10^{-6}}
\newcommand{\olmoLmNbCuratedHoneBDz}{0.957}
\newcommand{\olmoLmNbCuratedHoneBDiff}{0.057}
\newcommand{\olmoLmNbCuratedHoneCPbonf}{2.3 \times 10^{-10}}
\newcommand{\olmoLmNbCuratedHoneCDz}{1.889}
\newcommand{\olmoLmNbCuratedHoneCDiff}{0.179}
\newcommand{\olmoLmNbCuratedHpaP}{0.028}
\newcommand{\olmoLmNbCuratedHpaDz}{0.375}
\newcommand{\olmoLmNbCuratedHpaDiff}{0.010}
\newcommand{\olmoLmNbCuratedN}{39}

%% file: results/rlhf_experiment/tables/cross_model_macros.tex
% Auto-generated by scripts/rlhf_experiment/5c_cross_model_table.py

\newcommand{\crossModelQwenFourB}{0.241}
\newcommand{\crossModelQwenEightB}{0.210}
\newcommand{\crossModelQwenTwoThreeFiveB}{0.209}
\newcommand{\crossModelLlamaThreeThreeSeventyB}{0.169}
\newcommand{\crossModelGptFiveNano}{0.260}
\newcommand{\crossModelGptFive}{0.262}
\newcommand{\crossModelClaudeSonnetFourFive}{0.171}
\newcommand{\crossModelOlmoBase}{0.485}
\newcommand{\crossModelOlmoSft}{0.330}
\newcommand{\crossModelOlmoDpo}{0.299}
\newcommand{\crossModelOlmoInstruct}{0.283}

%% file: results/rlhf_experiment/tables_length_matched/rlhf_diversity.tex
\begin{tabular}{lrrr}
\toprule
\multicolumn{4}{l}{\textbf{AlpacaEval}} \\
\midrule
Stage & $D_{Ca_n}$ mean & std & $n$ \\
\quad Base & 0.481 & 0.038 & 150 \\
\quad SFT & 0.329 & 0.084 & 150 \\
\quad DPO & 0.286 & 0.072 & 150 \\
\quad Instruct (RLVR) & 0.281 & 0.071 & 150 \\
\addlinespace
Contrast & $\Delta$ & $d_z$ & $p_{\mathrm{Bonf}}$ \\
\quad Base$>$SFT & $0.151$ & $1.615$ & $3.4 \times 10^{-24}$ \\
\quad SFT$>$DPO & $0.044$ & $0.675$ & $1.7 \times 10^{-13}$ \\
\quad Base$>$Instruct (RLVR) & $0.200$ & $2.425$ & $5.2 \times 10^{-26}$ \\
\quad DPO$\,\neq\,$Instruct\,(H1$'$) & $0.005$ & $0.227$ & $0.004^{*}$ \\
\midrule
\multicolumn{4}{l}{\textbf{NoveltyBench curated}} \\
\midrule
Stage & $D_{Ca_n}$ mean & std & $n$ \\
\quad Base & 0.481 & 0.030 & 39 \\
\quad SFT & 0.369 & 0.085 & 39 \\
\quad DPO & 0.312 & 0.082 & 39 \\
\quad Instruct (RLVR) & 0.303 & 0.086 & 39 \\
\addlinespace
Contrast & $\Delta$ & $d_z$ & $p_{\mathrm{Bonf}}$ \\
\quad Base$>$SFT & $0.112$ & $1.212$ & $4.1 \times 10^{-8}$ \\
\quad SFT$>$DPO & $0.057$ & $0.957$ & $2.8 \times 10^{-6}$ \\
\quad Base$>$Instruct (RLVR) & $0.179$ & $1.889$ & $2.3 \times 10^{-10}$ \\
\quad DPO$\,\neq\,$Instruct\,(H1$'$) & $0.010$ & $0.375$ & $0.028^{*}$ \\
\bottomrule
\end{tabular}

%% file: sections/08_limitations_workshop.tex
\section{Limitations}\label{sec:limitations}

We collect the main limitations of the framework and of our $D = C \times a_n$ choice in one place so readers can calibrate how strongly to update on the results above.
Several points below recap caveats already made in their natural context and cross-reference those for detail.

\paragraph{Measurement is relative to $\theta$'s perception.}
The metric evaluates diversity \emph{as perceived by the base model $\theta$}: if $\pi$'s outputs differ only in ways $\theta$ cannot distinguish from context, the metric underestimates diversity (Section~\ref{sec:motivation}, ``Important caveat'').
Concretely, the off-diagonal sign of the pairwise surprise-reduction matrix changes with $\theta$'s scale (Section~\ref{sec:cross-mode-learning}), so $a_k$ curves for the same response set are not directly comparable across different choices of $\theta$.

\paragraph{Metric selection saw the full Tevet benchmark.}
We chose $C \times a_n$ (rather than alternative summaries such as $C \times E$; Appendix~\ref{app:excess-entropy}) after observing its performance on the full Tevet diversity-eval benchmark, without holding out a validation split.
The Tevet numbers in Section~\ref{sec:tevet} may therefore be optimistically biased: the scalar form was selected with visibility into those results.
The OLMo-2-7B post-training experiment (Section~\ref{sec:rlhf-experiment}) was scored after the metric form was fixed and provides independent validation; the synthetic mode-count experiments (Section~\ref{sec:mode-count-scaling}) informally influenced the choice (had they failed, we would likely have revised the metric) and so are not strictly independent.

\paragraph{External benchmark performance trails the strongest baseline.}
On Tevet and Berant's diversity-eval benchmark, $C \times a_n$ trails SentBERT by \tevetCxAnVsSentBertOCAMinPct\%--\tevetCxAnVsSentBertOCAMaxPct\% OCA across the nine binary tasks (Section~\ref{sec:tevet}); scaling the base model to Qwen3-30B-A3B-Base \citep{yang2025qwen3} did not close the gap (Appendix~\ref{app:qwen3-comparison}).
We position ICL-based diversity metrics (of which $D_{Ca_n}$ is one instance we found to work well) as a principled complement to embedding-based metrics, not a replacement.

\paragraph{Length-matching drops short-response prompts.}
Reporting in bits/byte (rather than total bits) makes per-byte cross-entropy not invariant to response length, so the OLMo-2-7B experiment scores a length-matched subset and drops $111/300$ prompts whose common per-prompt byte length falls below $50$~bytes; full mechanics and the un-truncated robustness check are in Section~\ref{sec:exp-discussion}.

\paragraph{$D = C \times a_n$ is a pragmatic choice, not a theoretically derived one.}
The product form was selected because it captures the properties we wanted a diversity score to have (residual surprise remains after the base model has seen previous responses, low-coherence outputs are penalised, and the result stays on a bits-per-byte scale), not because it is derived from an axiomatic information-theoretic setup.
Other scalars respecting the same constraints (weighted integrals of $a_k - a_\infty$, slope-based scores, or coherence applied at a different point on the curve) could equally be defended (see ``The framework admits other metrics'' in Section~\ref{sec:exp-discussion}).
We adopt $D_{Ca_n}$ because it is interpretable and works empirically across our validation regimes, not because it is uniquely correct.
We invite other researchers to build on this work to develop other formulas that characterize the $a_k$ curve in meaningful ways.

%% file: sections/conclusion_workshop.tex
\section{Conclusion}\label{sec:conclusion}

We have presented a new approach to measuring diversity using in-context learning, of which $D_{Ca_n} = C \times a_n$ is the working instance we evaluate: a base model's in-context learning detects similarities between arbitrary numbers of responses in a single forward pass.
The approach requires no embedding model, no reference corpus, no human labels, and no special-purpose training.
The same pipeline scores AI samples and human-written sets.
On Tevet and Berant's human-grounded benchmark it is near the level of the strongest sentence-embedding baseline.
On the OLMo-2-7B post-training pipeline it drops across the base $\to$ SFT $\to$ DPO/RLVR stages, tracking the diversity loss attributed to RLHF-style training.

%% file: sections/impact_workshop.tex
\section*{Impact Statement}

This paper presents work which may be used in evaluation pipelines that select for more capable AI systems.
We view the prospect of increasingly capable future AI as one of unclear positive or negative net impact.
This work may also contribute to the automation of human creative roles.
We publish it in the hope that better understanding of machine-generated diversity will contribute to better evaluation and oversight, and that these benefits outweigh the risks.

%% file: sections/06_practical.tex
\section{Practical Considerations}\label{sec:practical}

\subsection{Formatting the Conditioning Context}\label{sec:formatting}

Computing $\theta(r_k \mid r_{<k}, p)$ requires feeding $\theta$ a context containing the prompt and previous responses.
The formatting choice affects results, and we use two formats depending on whether the responses are best read as parallel answers to the prompt or as continuations of it.
Both are implemented as \texttt{format\_conditioning\_context} in \texttt{src/icl\_diversity/core.py} and selected via a \texttt{format\_mode} argument throughout the pipeline.

\paragraph{Instruct format.} The default, used for the synthetic scenarios (Appendix~\ref{sec:scenario-validation}), the mode-count experiments (Appendix~\ref{sec:mode-count-scaling}), and the OLMo-2-7B post-training case study (Section~\ref{sec:rlhf-experiment}):

\begin{verbatim}
    [prompt p]

    Response A: [r_1]

    Response B: [r_2]

    Response C: [r_3]
\end{verbatim}

The prompt appears once and each response is introduced by a labelled header (``Response A:'', ``Response B:'', \ldots, with labels rolling over to ``AA'', ``AB'' past ``Z''), encouraging the base model's in-context learning to treat the responses as parallel answers to the same prompt.

\paragraph{Completion format.} Used for the Tevet and Berant evaluation (Section~\ref{sec:tevet}), where each response set is a continuation of a shared narrative prompt rather than an instruction-style answer:

\begin{verbatim}
    1. [prompt p] [r_1]

    2. [prompt p] [r_2]

    3. [prompt p] [r_3]
\end{verbatim}

The prompt is repeated immediately before each response so $\theta$ scores $r_k$ in the same prompt-as-context that produced it.
Per-token log-probabilities are extracted only over the response tokens; the bits accumulated over the repeated prompt prefix are excluded from $a_k$, keeping the bits-numerator and the per-byte denominator both attributable to $r_k$ alone.

\paragraph{Base vs.\ instruction-tuned models.} We use a base (non-instruction-tuned) model as $\theta$ for two reasons.
First, an instruction-tuned model has had its output distribution shaped by RLHF or similar procedures, so using one as $\theta$ reintroduces the kind of distributional bias we seek to avoid.
Second, an instruction-tuned $\theta$ may assign systematically different probabilities to text that follows or violates its alignment training, introducing a confound between coherence-as-fluency and coherence-as-alignment that the metric should not conflate.
Modern base models already exhibit strong in-context learning from pretraining alone, so this choice does not sacrifice ICL capability.
Optimizing the prompt format to maximally elicit in-context learning from $\theta$ is an interesting direction but out of scope for this paper; we use the neutral ``Response A/B/C'' template throughout.

\subsection{Computational Cost}\label{sec:cost}

Because $\theta$ is a causal language model, the full $a_k$ curve for one response ordering can be computed from a \textbf{single forward pass}.
The prompt and all responses are concatenated into one sequence (Figure~\ref{fig:single-pass-layout}).
The forward pass produces the log-probability of every token conditioned on all preceding tokens.

\begin{figure*}[t]
\centering
\begin{verbatim}
[prompt p] Response A: [r_1] Response B: [r_2] ... Response N: [r_n]
\end{verbatim}
\caption{Single-pass input layout for computing the full $a_k$ curve.
The prompt is followed by all $n$ responses with ``Response A/B/C/\ldots'' labels; one forward pass over this sequence yields every $a_k$ value simultaneously, since causal attention conditions the tokens of $r_k$ on exactly $p, r_1, \ldots, r_{k-1}$ (plus formatting).}
\label{fig:single-pass-layout}
\end{figure*}
The tokens belonging to $r_k$ are conditioned on exactly $p, r_1, \ldots, r_{k-1}$ (plus formatting), which is the conditioning required for $a_k$.
Partitioning the output log-probabilities by response boundary and summing within each partition yields all $n$ values of $a_k$ simultaneously for that ordering, with FLOPs scaling as $O((|p| + n\bar{L}_{\mathrm{tok}})^2)$ from causal attention, where $\bar{L}_{\mathrm{tok}}$ is the average response length in tokens.
Because we report $\bar{a}_k$ averaged over $|\Sigma|$ random permutations of the response ordering (Section~\ref{sec:ordering}), the total $a_k$-curve cost is $|\Sigma|$ such passes; the permutations are independent and we batch them on a single GPU.

The coherence term $C$ additionally requires the \emph{unconditional} per-byte cross-entropies $h_\theta(r_i \mid p)$ for each response.
These cannot be extracted from the concatenated pass, since in that pass $r_i$ for $i > 1$ is conditioned on all prior responses.
Computing them requires $n$ independent forward passes, each over the short context $(p, r_i)$.
These passes are embarrassingly parallel and batchable, and they do not depend on the ordering, so $C$ is computed once per response set regardless of $|\Sigma|$.

In summary, scoring one (prompt, response set) tuple takes:
\begin{itemize}[nosep]
    \item $a_k$ curve: $|\Sigma|$ forward passes (long context, one per permutation).
    \item $C$: $n$ forward passes (short contexts, batchable; not multiplied by $|\Sigma|$).
\end{itemize}
With $n = 10$ responses and $|\Sigma| = 50$ permutations (our default for larger-scale experiments), this is $50 + 10 = 60$ forward passes per (prompt, response set) tuple; the long-context $|\Sigma|$ passes dominate wall-clock time.

\subsection{Dependence on Sample Ordering}\label{sec:ordering}

The $a_k$ values depend on the ordering of $\{r_i\}$.
Individual responses differ in how surprising they are to $\theta$ given the responses that precede them, so $a_k$ depends on which response sits at position $k$ and which others precede it, making a single ordering's curve jagged.
Averaging over random permutations removes this dependence: each position averages over many choices of response and preceding context, so the curve reflects only how $\theta$'s predictions improve with more context.
We therefore average over $|\Sigma|$ random permutations; Section~\ref{sec:practical-findings} compares low- and high-permutation runs on the validation scenarios and finds that the low setting misranks adjacent scenarios.
We have not swept intermediate values and cannot pinpoint a cheap-and-reliable minimum, so we default to $|\Sigma| = 100$ for scenario-level experiments and $|\Sigma| = 50$ elsewhere.

Per-response normalization by byte count must be performed \emph{before} averaging across permutations, since the byte count depends on which response lands at each position.
Concretely, the per-byte curve is a mean of per-permutation rates,
\begin{equation}\label{eq:akbar-mor}
    \bar{a}_k = \frac{1}{|\Sigma|}\sum_{\sigma \in \Sigma} \frac{a_k^{\sigma}}{\|r_{\sigma(k)}\|},
\end{equation}
not a ratio of permutation-averaged bits to permutation-averaged byte counts (which would degenerate, after enough permutations, into the total-bits curve rescaled by the mean response length).

\subsection{Choice of $n$}

The diversity score depends on $n$ as a measurement parameter, not through any estimate of an asymptotic floor.
Given enough in-context examples, even genuinely diverse policies eventually become predictable to $\theta$, so $a_k$ keeps decreasing rather than converging to a meaningful irreducible value: there is no $a_\infty$ for $a_n$ to approximate.
Too small an $n$ means $\theta$ has not yet exploited the inter-response structure, and $D_{Ca_n}$ overstates diversity.
Too large an $n$ means $\theta$ has accumulated enough in-context examples to reduce its surprise even within genuine diversity; for any policy with finitely many distinct modes, $a_n$ continues to decrease toward zero as $\theta$ learns to predict within-mode variation from prior examples.
For comparisons to be meaningful, $n$ must be held fixed across all policies under evaluation.
Our experiments use $n = 5$ on Tevet (Section~\ref{sec:tevet}), $n = 10$ on OLMo-2-7B and scenario validation, and $n = 20$ on the mode-count experiment (Appendix~\ref{sec:mode-count-scaling}).
This is one reason to report the full $a_k$ curve rather than $D_{Ca_n}$ alone. $D$ is a lossy summary, but the curve shows directly how response surprise evolves at every context size.

%%%%%%%%%%%%%%%%%%%%%%%%%%%%%%%%%%%%%%%%%%%%%%%%%%%%%%%%%%%%%%%%%%%%%%%%
%% SECTION 8: EXPERIMENTS
%%%%%%%%%%%%%%%%%%%%%%%%%%%%%%%%%%%%%%%%%%%%%%%%%%%%%%%%%%%%%%%%%%%%%%%%

%% file: sections/07_experiments_intro.tex
\section{Experiments}\label{sec:experiments}

\subsection{Experimental Setup}\label{sec:exp-setup}

We evaluate the metric primarily using Qwen2.5-3B (3B parameters, 32K-token context window) \citep{yang2024qwen25}, with GPT-2 (124M parameters, 1024-token context window) \citep{radford2019language} as a smaller-model comparison point for scenario validation.
Qwen2.5-3B is used for mode-count scaling, cross-mode learning, and external validation (Sections~\ref{sec:mode-count-scaling}, \ref{sec:cross-mode-learning}, \ref{sec:tevet}).
The cross-model scaling study (Section~\ref{sec:exp-discussion}) additionally uses the full Llama 3 family for comparison.

All $a_k$ curves are computed using single-pass forward computation (Section~\ref{sec:cost}).
Unless otherwise noted, we use 100 permutations for scenario validation and 50 permutations for larger-scale experiments.
Responses are formatted as described in Section~\ref{sec:cost}.
All code and data, including the Olmo-2 RLHF generations under \texttt{results/rlhf\_experiment/}, are publicly available.\footnote{\projectGithubUrl}

%% file: sections/07_2_scenario_validation.tex
\subsection{Scenario Validation}\label{sec:scenario-validation}

We construct five synthetic scenarios with known diversity structure:
\begin{enumerate}[nosep]
    \item \textbf{Pure noise}: Random ASCII characters (letters, digits, punctuation, spaces) with no learnable structure.
    \item \textbf{Multi-incoherent}: Responses from 5 distinct modes, each internally incoherent (scrambled words within a template).
    \item \textbf{Multi-mode}: Responses from 5 distinct modes, each internally coherent (e.g., recipe, poem, code).
    \item \textbf{One-mode}: All responses from a single coherent mode (paraphrases of the same content).
    \item \textbf{Mixed}: A mixture of coherent and incoherent responses.
\end{enumerate}

Each scenario contains 10 responses per prompt, 5 prompts per scenario, with 100 permutations and seed 42.

\begin{table*}[t]
\centering
\caption{Scenario validation metrics (mean across 5 prompts, 100 permutations, $n=10$ responses, per-byte).
We report several candidate scalars: unconditional surprise $a_1$, the curve's last point $a_n$, coherence $C$, our recommended score $D_{Ca_n} = C \times a_n$ (\textbf{bold}), the excess entropy $E$ (Appendix~\ref{app:excess-entropy}), $C \times E$, and the coherence spread $\sigma_\ell$ (the standard deviation of the per-byte cross-entropies $\{h_\theta(r_i \mid p)\}_{i=1}^n$). $D_{Ca_n}$ correctly ranks multi-mode above one-mode and suppresses incoherent scenarios via $C$.
Both $E$ and $C \times E$ go negative for the mixed scenario on GPT-2.
Mixed empirically scores the highest $D_{Ca_n}$ on both models, exceeding multi-mode coherent rather than landing at the ``mid'' position Table~\ref{tab:edge-cases} predicts; this is not the intended ranking.
The coherence spread $\sigma_\ell$ flags this case in its intended diagnostic role (the largest $\sigma_\ell$ across scenarios on both models), reflecting the within-set heterogeneity between coherent and incoherent responses; reweighting variants such as $C^{\alpha} \times a_n$ are discussed in Section~\ref{sec:exp-discussion}.}
\label{tab:scenarios}
\small
\setlength{\tabcolsep}{2pt}
\input{results/tables/scenario_validation.tex}
\setlength{\tabcolsep}{6pt}
\end{table*}

\begin{figure*}[t]
\centering
\includegraphics[width=\textwidth,height=0.4\textheight,keepaspectratio]{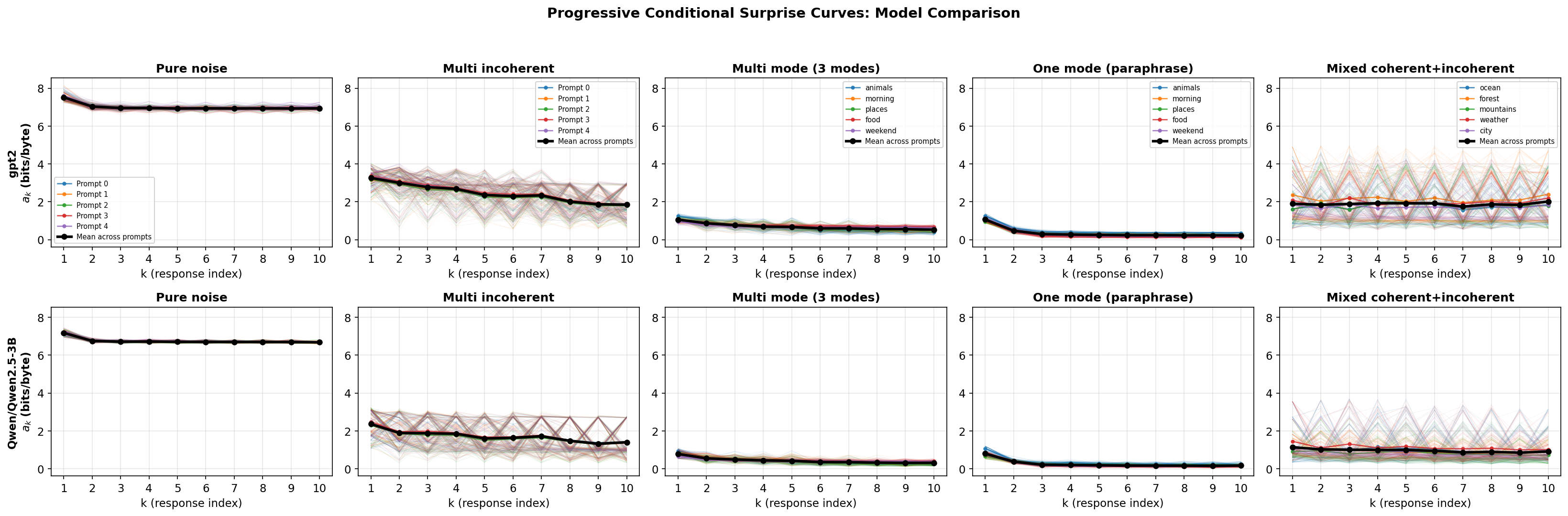}
\caption{Progressive conditional surprise curves ($a_k$, per-byte) for all five scenarios, comparing GPT-2 (top row) and Qwen2.5-3B (bottom row).
In each panel, faint colored curves are individual permutations (100 per prompt), medium colored curves are per-prompt averages, and the bold black curve is the mean across prompts.
Pure noise curves are flat (no learnable structure); multi-mode curves show progressive decline; one-mode curves decline steeply then flatten.}
\label{fig:scenario-curves}
\end{figure*}

\paragraph{Results.} Table~\ref{tab:scenarios} summarizes the metrics.
Figure~\ref{fig:scenario-curves} shows the $a_k$ curves for all scenarios.

The coherence term $C$ correctly separates coherent from incoherent scenarios: $C \approx 0.5$--$0.6$ for multi-mode and one-mode, $C \approx 0.1$--$0.3$ for multi-incoherent, and $C \approx 0.01$ for pure noise.
The coherence spread $\sigma_\ell$ discriminates mixed from pure scenarios: $\sigma_\ell > 1.0$ for mixed (GPT-2), reflecting the within-set heterogeneity between coherent and incoherent responses.

$C$ consistently improves with model strength (Qwen2.5-3B assigns higher coherence to coherent text than GPT-2). $D_{Ca_n}$ correctly ranks multi-mode above one-mode on both models, and suppresses pure noise and multi-incoherent via the $C$ factor.

%% file: results/tables/scenario_validation.tex
% Generated by: scripts/generate_scenario_table.py
% Data sources: results/scenario_metrics_v3_gpt2_100perm.json
%               results/scenario_metrics_v3_qwen3b_100perm.json
\begin{tabular}{@{}l rrrrrrr rrrrrrr@{}}
\toprule
& \multicolumn{7}{c}{\textbf{GPT-2 (124M)}} & \multicolumn{7}{c}{\textbf{Qwen2.5-3B}} \\
\cmidrule(lr){2-8} \cmidrule(lr){9-15}
Scenario & $a_1$ & $a_n$ & $C$ & $\mathbf{D_{Ca_n}}$ & $E$ & $C\!\times\!E$ & $\sigma_\ell$ & $a_1$ & $a_n$ & $C$ & $\mathbf{D_{Ca_n}}$ & $E$ & $C\!\times\!E$ & $\sigma_\ell$ \\
\midrule
Pure noise & 7.52 & 6.94 & 0.005 & \textbf{0.04} & 0.68 & 0.004 & 0.20 & 7.18 & 6.68 & 0.007 & \textbf{0.05} & 0.75 & 0.005 & 0.11 \\
Multi-incoher. & 3.26 & 1.85 & 0.10 & \textbf{0.19} & 5.94 & 0.61 & 0.48 & 2.36 & 1.40 & 0.21 & \textbf{0.29} & 3.12 & 0.64 & 0.69 \\
Multi-mode & 1.06 & 0.53 & 0.48 & \textbf{0.26} & 1.61 & 0.76 & 0.08 & 0.78 & 0.31 & 0.59 & \textbf{0.19} & 1.17 & 0.68 & 0.11 \\
One-mode & 1.06 & 0.22 & 0.48 & \textbf{0.10} & 1.31 & 0.63 & 0.06 & 0.80 & 0.16 & 0.58 & \textbf{0.09} & 0.92 & 0.53 & 0.08 \\
Mixed & 1.90 & 2.02 & 0.26 & \textbf{0.52} & $-$1.29 & $-$0.31 & 1.17 & 1.13 & 0.91 & 0.46 & \textbf{0.41} & 0.52 & 0.23 & 0.73 \\
\bottomrule
\end{tabular}

%% file: sections/07_3_mode_count.tex
\subsection{Mode Count Scaling}\label{sec:mode-count-scaling}

To test whether the metric reflects the number of distinct modes, we construct response sets with $m \in \{1, \ldots, \modeCountMMax\}$ modes drawn from a pool of \modeCountPoolSize{} format-based generators (haiku, code, recipe, legal disclaimer, etc.), all responding to the same prompt (``Write a short piece about rain'').
For Qwen2.5-3B: $n=\modeCountNResponses$ responses, \modeCountNDraws{} random draws of mode assignments, averaged over permutations.

\begin{figure}[!htbp]
\centering
\includegraphics[width=\columnwidth]{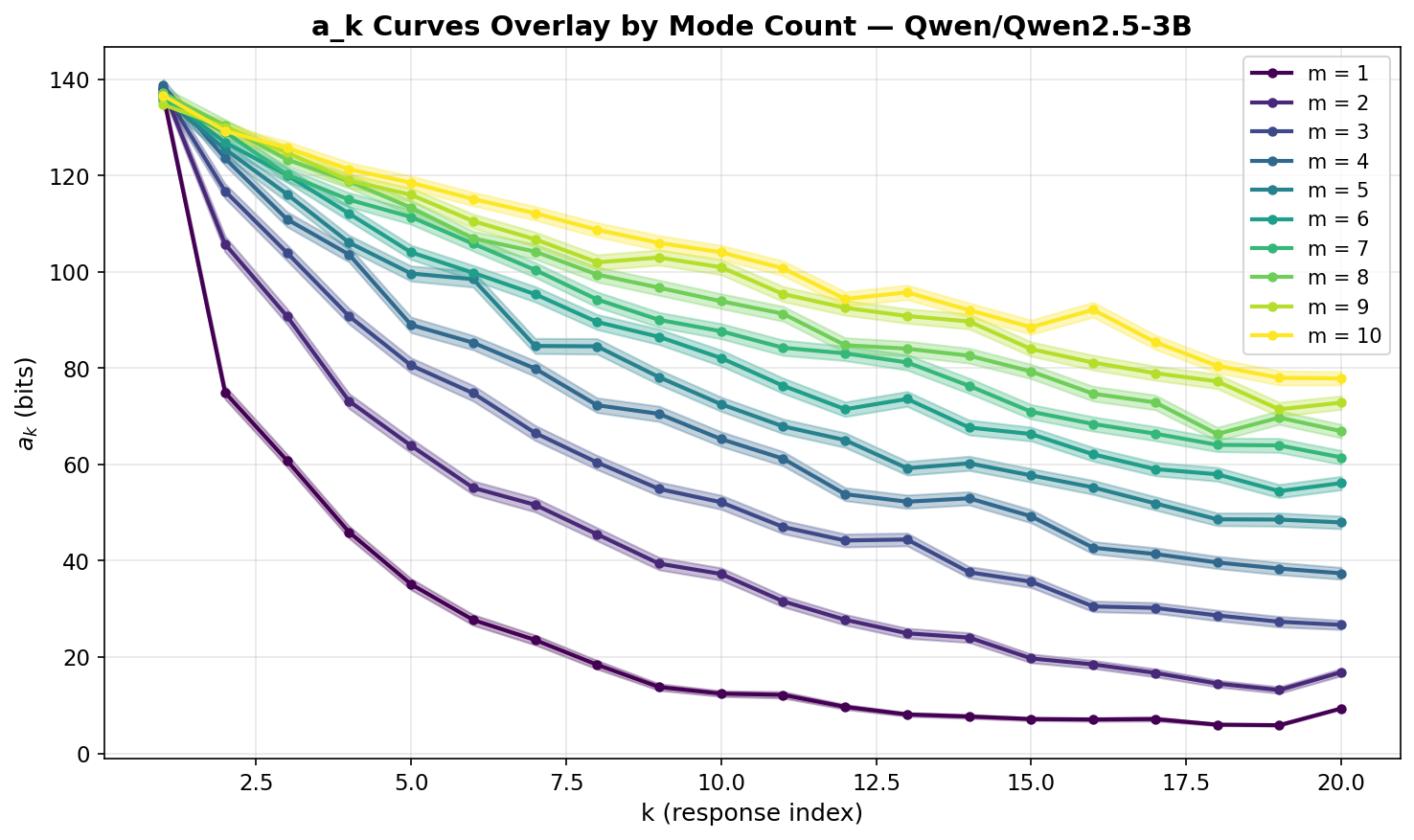}
\caption{Mode count scaling on Qwen2.5-3B ($n=\modeCountNResponses$, \modeCountNDraws{} draws).
The $a_k$ curves fan out with increasing $m$: higher floors, slower convergence.
All curves are exponential (no sigmoidal plateau), even at $m=\modeCountMMax$, due to cross-mode learning (Section~\ref{sec:cross-mode-learning}).
Shaded bands around each curve are $\pm 1$ standard error of the mean across the \modeCountNDraws{} random draws of mode assignments (i.e., the across-draw standard deviation divided by $\sqrt{\modeCountNDraws}$); they are narrow because averaging over \modeCountNDraws{} draws sharply reduces the across-draw spread.}
\label{fig:mode-count}
\end{figure}

\paragraph{Key findings.} Figure~\ref{fig:mode-count} shows the $a_k$ curves.
$a_n$ increases monotonically with $m$ (\modeCountAnMone{} bits at $m=1$ to \modeCountAnMten{} bits at $m=10$; see Table~\ref{tab:mode-count}), reflecting that with more modes there are fewer same-mode repetitions within $n$ responses, so each response remains more surprising even after $\theta$ has seen the others.
This is exactly the behavior $D_{Ca_n} = C \times a_n$ needs: $a_n$ tracks mode count directly.
All curves are purely exponential (no sigmoidal plateau), even at $m=10$; Section~\ref{sec:cross-mode-learning} investigates why.
For a detailed comparison of the excess-entropy-based scores on this data, including the sigmoid fit parameters and the non-monotonicity of $\hat{E}_n$, see Appendix~\ref{app:excess-entropy}.

%% file: sections/07_4_cross_mode.tex
\subsection{Cross-Mode Learning and Curve Shape}\label{sec:cross-mode-learning}

The authors predicted that at high $m$, the $a_k$ curve should be sigmoidal: an initial plateau (early responses come from different modes and are uninformative about each other), followed by decline once mode repetitions accumulate.
GPT-2 shows hints of this pattern at some $m$ values (e.g. $m=10$).
Qwen2.5-3B does not: it shows immediate exponential decay at all $m$.
This section investigates the discrepancy through systematic pairwise analysis.

\paragraph{Pairwise cross-mode surprise matrix.} For 15 modes, we compute a $15 \times 15$ matrix $M_{ij}$: the surprise reduction (in bits) when a response from mode $j$ is in the context and mode $i$ is the target.
Diagonal entries measure same-mode learning; off-diagonal entries measure cross-mode information transfer.
Each entry averages over 5 context--target samples (using different samples for context and target to avoid self-prediction inflation).

\begin{figure*}[p]
\centering
\begin{subfigure}{\textwidth}
    \centering
    \includegraphics[width=\textwidth]{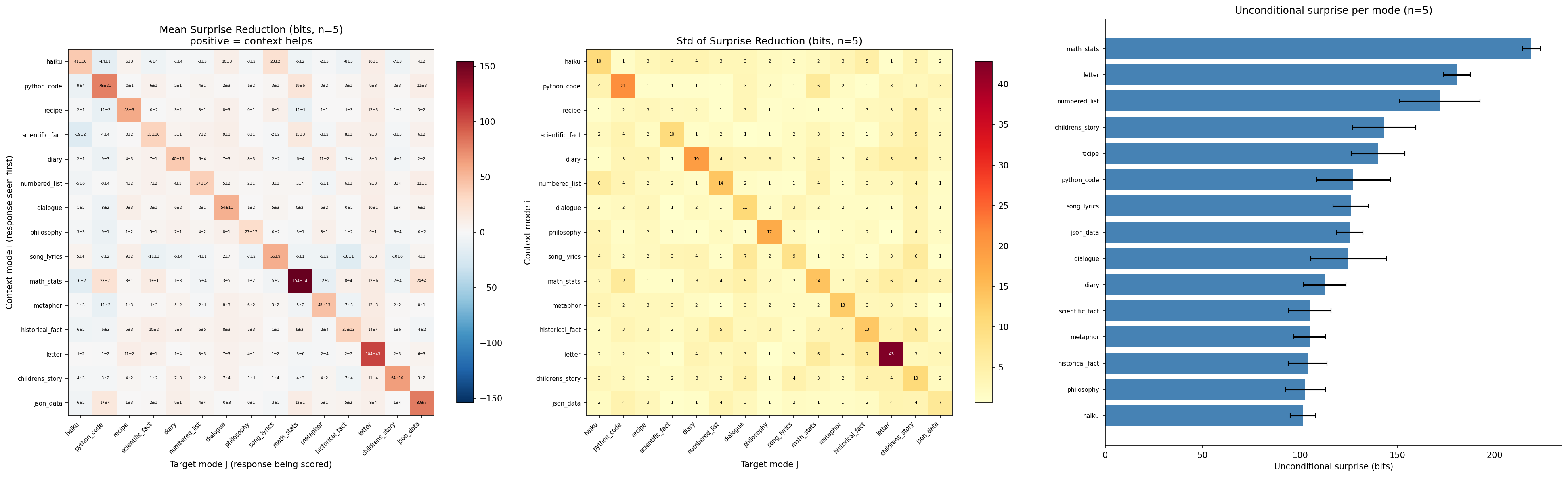}
    \caption{Qwen2.5-3B: positive off-diagonal.}
    \label{fig:pairwise-qwen}
\end{subfigure}

\vspace{1.5em}

\begin{subfigure}{\textwidth}
    \centering
    \includegraphics[width=\textwidth]{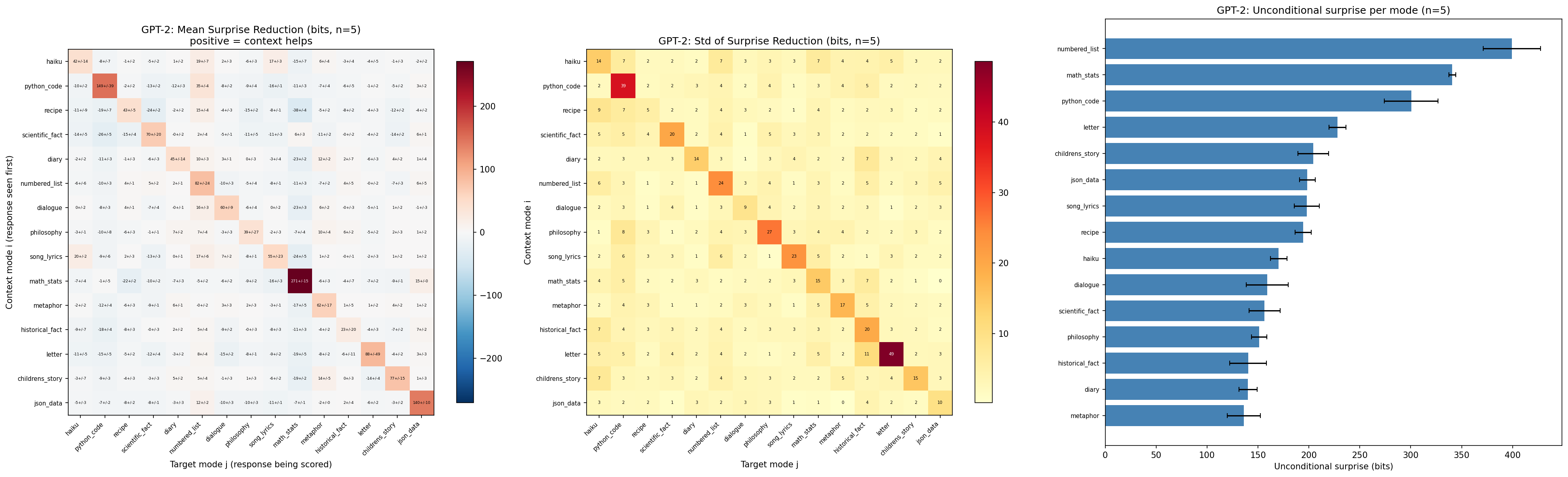}
    \caption{GPT-2: negative off-diagonal.}
    \label{fig:pairwise-gpt2}
\end{subfigure}
\caption{Pairwise cross-mode surprise reduction matrices.
Each cell $(i,j)$ shows how many bits of surprise reduction mode $i$ (target) receives from seeing mode $j$ (context).
Qwen shows diagonal dominance with pervasive positive off-diagonal (\crossmodeQwenOffMean{} bits mean); GPT-2 shows diagonal dominance with negative off-diagonal (\crossmodeGPTOffMean{} bits mean).}
\label{fig:pairwise-matrices}
\end{figure*}

\paragraph{Results.} (Cell standard deviations are the mean within-cell sampling std across the \crossmodeNumModes{} diagonal or \crossmodeNumOffPairs{} off-diagonal matrix cells, each estimated from 5 context-target sample pairs.)
\begin{itemize}[nosep]
    \item \textbf{Qwen2.5-3B}: diagonal mean = $\crossmodeQwenDiagMean \pm \crossmodeQwenDiagStd$ bits, off-diagonal mean = $\crossmodeQwenOffMean \pm \crossmodeQwenOffStd$ bits, \crossmodeQwenOffPctPos\% of off-diagonal entries positive.
    \item \textbf{GPT-2}: diagonal mean = $\crossmodeGPTDiagMean \pm \crossmodeGPTDiagStd$ bits, off-diagonal mean = $\crossmodeGPTOffMean$ bits, only \crossmodeGPTOffPctPos\% of off-diagonal entries positive.
\end{itemize}

The sign of the off-diagonal determines the curve shape (Figure~\ref{fig:pairwise-matrices}):
\begin{itemize}[nosep]
    \item \textbf{Positive off-diagonal (Qwen)}: Every response in context lowers expected surprise on every other response.
The $a_k$ curve drops from position 1, producing exponential decay.
    \item \textbf{Negative off-diagonal (GPT-2)}: Cross-mode responses actively raise surprise on subsequent responses.
At $m=10$, the expected first-step net drop under the additive independent-mode model is only $\crossmodeGPTFirstStepDrop$ bits (the \crossmodeGPTDiagGainMten{}-bit diagonal gain from the $1/m$ same-mode chance is largely offset by the $\crossmodeGPTCrossDamageMten$-bit cross-mode damage from the $(m{-}1)/m$ different-mode chance), producing a flat plateau until same-mode repetitions accumulate, the predicted sigmoid.
\end{itemize}

GPT-2 has a larger diagonal and a negative off-diagonal: it gains more from same-mode context but is actively penalised by cross-mode context.
The two effects are linked, since probability mass that conditioning concentrates on same-mode continuations is mass taken away from other modes; a sharper conditional on the seen mode is necessarily worse on the unseen ones.
Qwen does the opposite: a smaller diagonal but a positive off-diagonal, distributing its update across modes.

\begin{figure}[!htbp]
\centering
\includegraphics[width=\columnwidth]{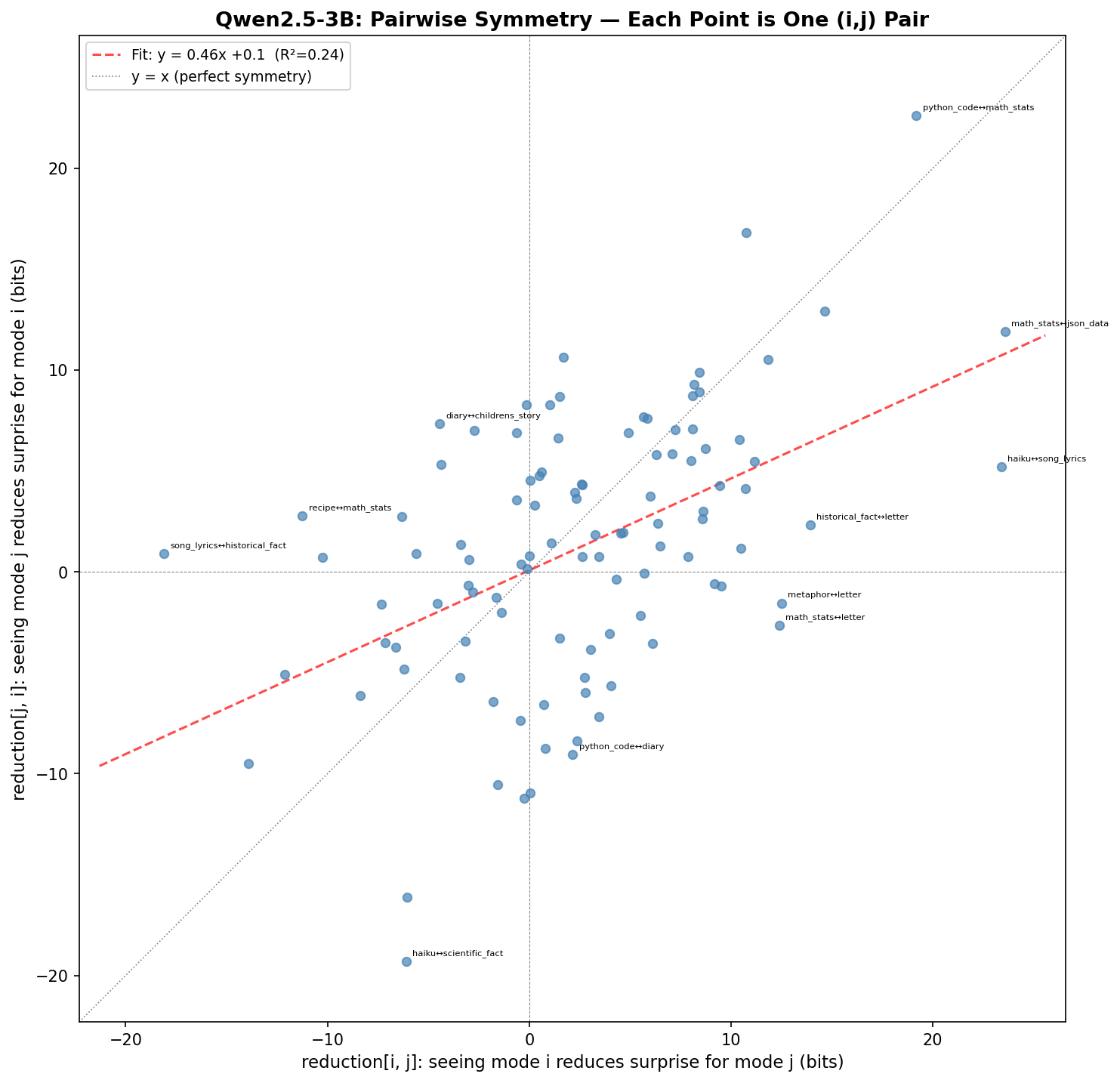}
\caption{Pairwise symmetry scatter for Qwen2.5-3B: each point is one $(i,j)$ pair, plotting $M_{ij}$ vs.\ $M_{ji}$.
Under symmetric mutual information, points would lie on $y = x$.
The best-fit slope is \scalingSymSlopeQwen{} ($R^2 = \scalingSymRsqQwen$), and many pairs have opposite signs: mode $i$ can reduce surprise for mode $j$ while $j$ \emph{increases} surprise for $i$.}
\label{fig:pairwise-symmetry}
\end{figure}

\paragraph{Pairwise asymmetry.} The mean absolute asymmetry $|M_{ij} - M_{ji}|$ is \crossmodeAsymmetryMean{} bits ($\crossmodeAsymmetryRatio\times$ the off-diagonal mean), falsifying the hypothesis that pairwise surprise reduction is symmetric (Figure~\ref{fig:pairwise-symmetry}).
Information-theoretically $I(X;Y) = I(Y;X)$, so this asymmetry cannot reflect asymmetric information structure between modes; it reflects $\theta$'s imperfection as an in-context reasoner. The same content is more useful as context than as target depending on its surface form, mirroring the \emph{reversal curse} observed in autoregressive LLMs trained on ``A is B'' but failing on ``B is A'' \citep{berglund2023reversal}.

\paragraph{Connections to information-theoretic properties.} Two information-theoretic properties are relevant to interpreting the pairwise matrix:

\begin{enumerate}[nosep]
    \item \textbf{Non-negative conditioning} ($H(X) \geq H(X \mid Y)$): Under the true distribution, conditioning can never increase entropy on average.
For cross-entropies this is not guaranteed (the KL term can increase), but a well-calibrated $\theta$ should show mostly non-negative off-diagonal entries.
    \item \textbf{Symmetry of mutual information} ($I(X;Y) = I(Y;X)$): Under the true joint, row means (how informative mode $i$ is as context) should correlate with column means (how much mode $i$ benefits from context).
\end{enumerate}

\begin{figure*}[t]
\centering
\begin{subfigure}[t]{0.48\textwidth}
    \includegraphics[width=\textwidth]{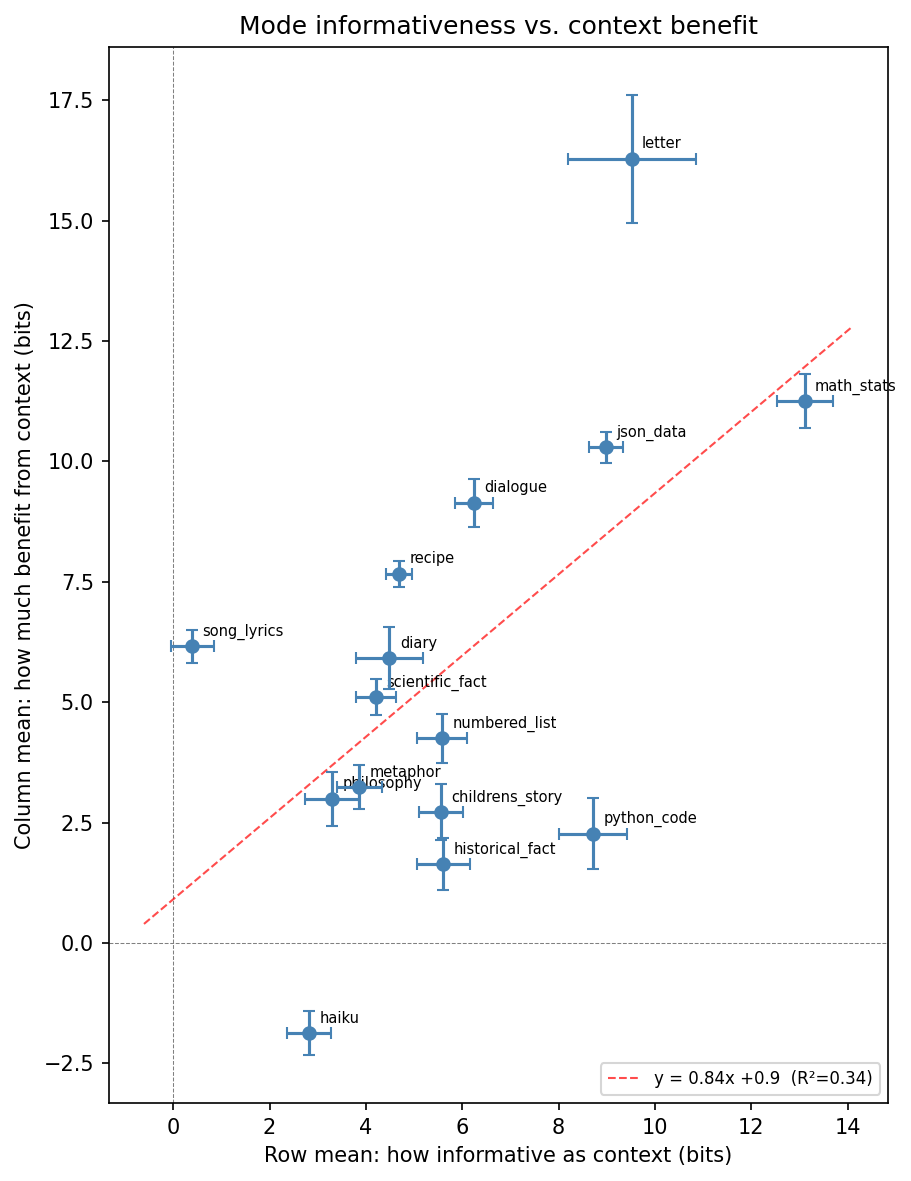}
    \caption{Qwen2.5-3B: slope = \scalingRCSlopeQwen{}, $R^2 = \scalingRCRsqQwen$.}
    \label{fig:rowcol-qwen}
\end{subfigure}
\hfill
\begin{subfigure}[t]{0.48\textwidth}
    \includegraphics[width=\textwidth]{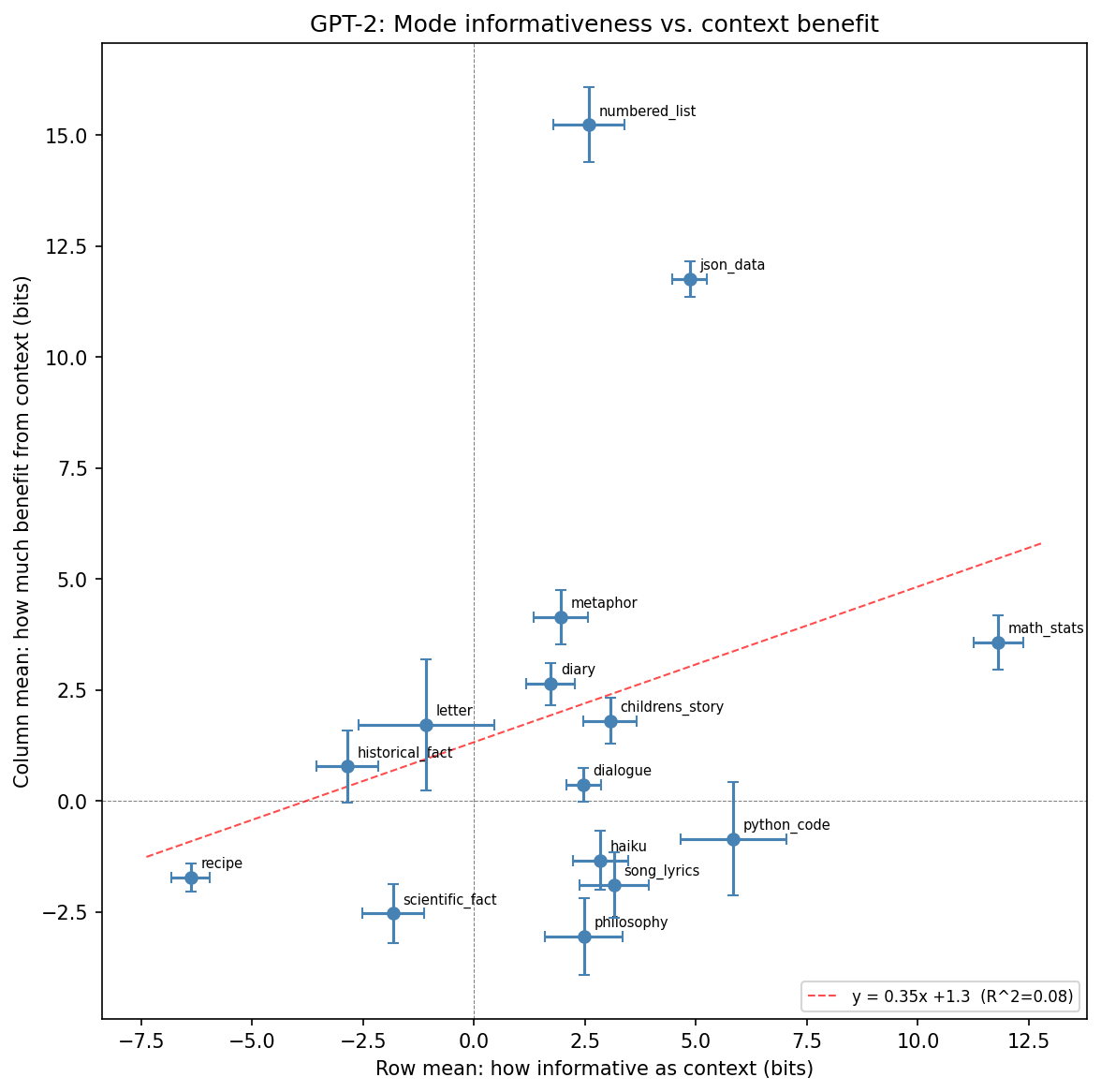}
    \caption{GPT-2: slope = \scalingRCSlopeGPT{}, $R^2 = \scalingRCRsqGPT$.}
    \label{fig:rowcol-gpt2}
\end{subfigure}
\caption{Row mean (how informative as context) vs.\ column mean (how much benefit from context) for each mode.
Qwen shows tighter correlation closer to the identity, consistent with approximate symmetry of mutual information.
GPT-2 shows widespread violations of non-negative conditioning (modes in the negative-row-mean region).}
\label{fig:row-vs-col}
\end{figure*}

Figure~\ref{fig:row-vs-col} compares the two models.
Qwen shows a tighter row--column relationship (slope = \scalingRCSlopeQwen{}, $R^2 = \scalingRCRsqQwen$) with only \crossmodeQwenNegColCount{} of \crossmodeNumModes{} modes having a negative column mean.
GPT-2 shows widespread violations of non-negative conditioning (\crossmodeGPTNegColCount{} modes with negative column means) and a barely-existent row--column relationship (slope = \scalingRCSlopeGPT{}, $R^2 = \scalingRCRsqGPT$).
Across the four Llama models \citep{grattafiori2024llama3,meta2024llama32} we additionally tested (Section~\ref{sec:exp-discussion}, Figure~\ref{fig:scaling-crossmode}), the fraction of positive off-diagonal entries rises monotonically with size (\scalingFracPosLlamaOneB\% at 1B to \scalingFracPosLlamaSeventyB\% at 70B), and the off-diagonal mean follows the same trend (\scalingOffDiagLlamaOneB{} bits at 1B to \scalingOffDiagLlamaSeventyB{} bits at 70B): the \emph{magnitude} of cross-mode information transfer increases with size.
The row--column relationship does not: $R^2$ peaks at Llama-3B (\scalingRCRsqLlamaThreeB{}) and degrades at the larger Llamas (\scalingRCRsqLlamaSeventyB{} at 70B), so the \emph{symmetry} of the pairwise matrix does not correspondingly improve.

\begin{figure*}[!htbp]
\centering
\includegraphics[width=\textwidth]{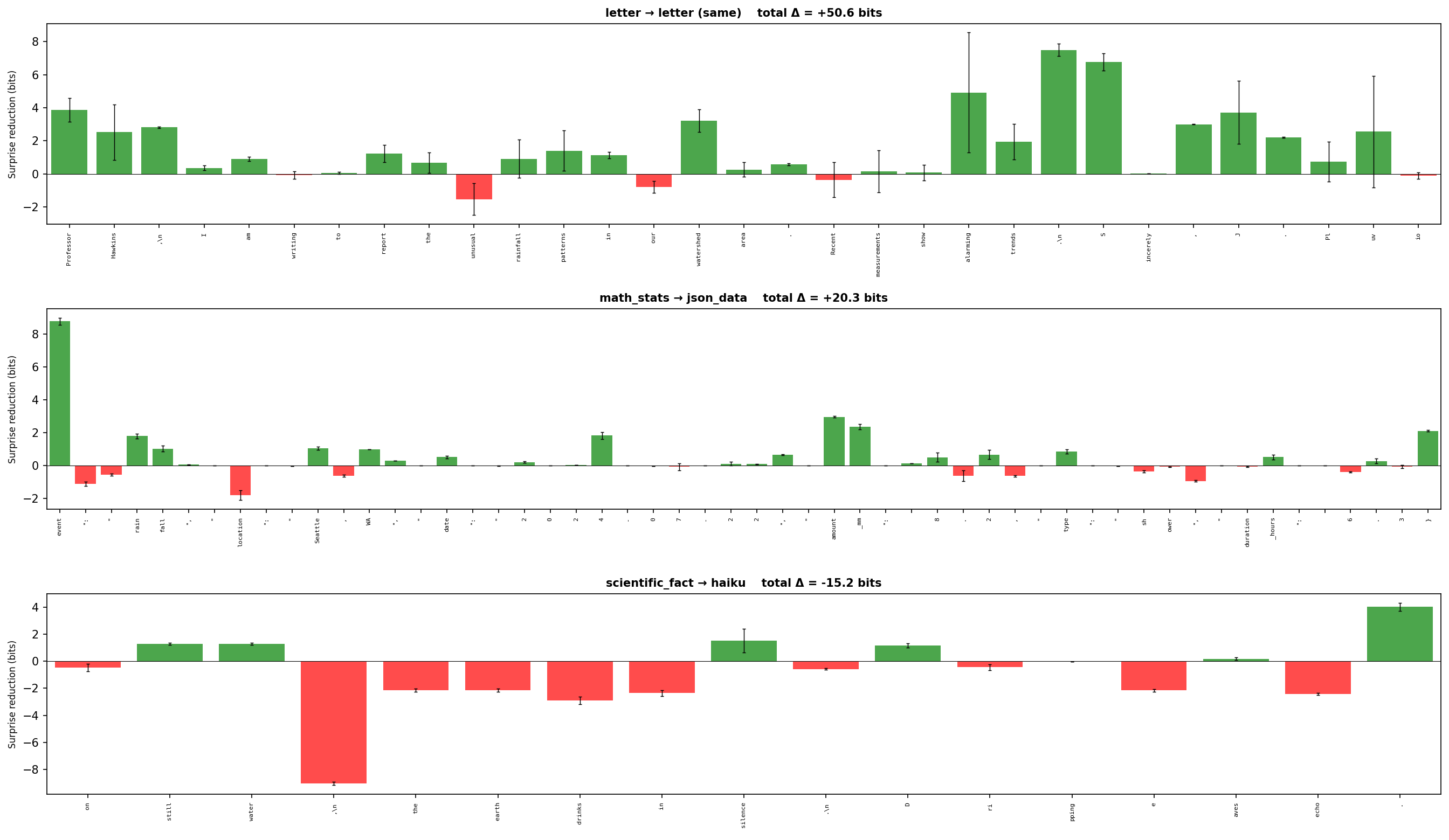}
\caption{Per-token surprise reduction (bits) for three representative context--target pairs (Qwen2.5-3B). Bars are mean reductions across $5$ context samples per pair; error bars are $\pm 1$ standard deviation. Green bars indicate the conditioning context lowered the per-token surprise; red bars indicate it raised it. The three pairs are stratified-median selections (by total $|\Delta|$) from the same-mode, positive cross-mode, and negative cross-mode strata of a $9$-pair pool. Note how in scientific fact to haiku, the early newline triggers the largest surprise. Scientific facts may discuss "still water", but they almost never have a line with three words in it.}
\label{fig:token-attribution}
\end{figure*}

\paragraph{Token-level attribution.} Figure~\ref{fig:token-attribution} shows per-token surprise reduction (in bits) for one pair from each of three strata: same-mode (\texttt{letter $\to$ letter}), cross-mode where conditioning helps (\texttt{math\_stats $\to$ json\_data}), and cross-mode where conditioning hurts (\texttt{scientific\_fact $\to$ haiku}). Pairs were selected by stratified-median total $|\Delta|$.

\paragraph{Discussion.} The paper's sigmoid prediction assumes modes are approximately independent.
For capable models, this fails: cross-mode learning creates positive information transfer between distinct modes.
The Llama 1B/3B/8B/70B series confirms this scales with model size (Section~\ref{sec:exp-discussion}): the total cross-mode information $\theta$ extracts grows monotonically with size, and with it non-negative conditioning is better satisfied.
The matrix nevertheless remains asymmetric (row--column $R^2$ peaks at Llama-3B and degrades at the larger Llamas), which under $I(X;Y) = I(Y;X)$ signals that even Llama-70B is an imperfect in-context reasoner.

%% file: sections/07_7_practical_findings.tex
\subsection{Practical Findings}\label{sec:practical-findings}

\paragraph{Permutation sensitivity.} The per-byte $a_k$ curve depends on the response ordering.
Comparing the $D_{Ca_n}$ ranking of the \permSensNumScenarios{} validation scenarios from Section~\ref{sec:scenario-validation} between \permSensLowPerm{}- and \permSensHighPerm{}-permutation runs: on GPT-2, \permSensGPTMisranked{} of \permSensNumScenarios{} scenarios shift rank between the two settings; on Qwen2.5-3B, \permSensQwenMisranked{} of \permSensNumScenarios{}.
Where scenarios do shift rank, it is between adjacent scenarios with similar $D_{Ca_n}$ values rather than dramatic reorderings.
We use \permSensHighPerm{} permutations throughout for scenario-level experiments; we have not swept intermediate values, so we cannot pinpoint a minimum that is both cheap and reliable.
The coherence term $C$ is permutation-invariant by construction and unaffected.

\paragraph{Boundary handling.} When Qwen merges a response's final \texttt{.} with the following \texttt{\textbackslash n\textbackslash n} into a single \texttt{.\textbackslash n\textbackslash n} token, our boundary detector (which uses a character-span overlap rule) attributes that token to the response, not the separator.
The response's cross-entropy therefore includes a small contribution from predicting the upcoming separator alongside its own characters; the byte-count denominator remains the response's literal byte length.
We have not measured the magnitude of this bias on our reported numbers.
Boundary correctness is verified by \texttt{tests/test\_response\_boundaries.py}.

%% file: sections/07_8_discussion.tex
\subsection{Discussion}\label{sec:exp-discussion}

\paragraph{$\sigma_\ell$ is a diagnostic, not a diversity signal.} $\sigma_\ell$ increased monotonically with $m$ in our synthetic experiments because our modes have very different formats (code vs.\ haiku vs.\ recipe $\to$ different per-byte cross-entropies).
But $\sigma_\ell$ measures coherence \emph{heterogeneity}, not diversity: modes with similar fluency would have low $\sigma_\ell$ despite high diversity, and a single mode with high quality variance could have high $\sigma_\ell$ with zero diversity.
Its role is diagnostic (detecting mixed coherence), not as a standalone diversity score.

\begin{figure*}[t]
\centering
\includegraphics[width=\textwidth]{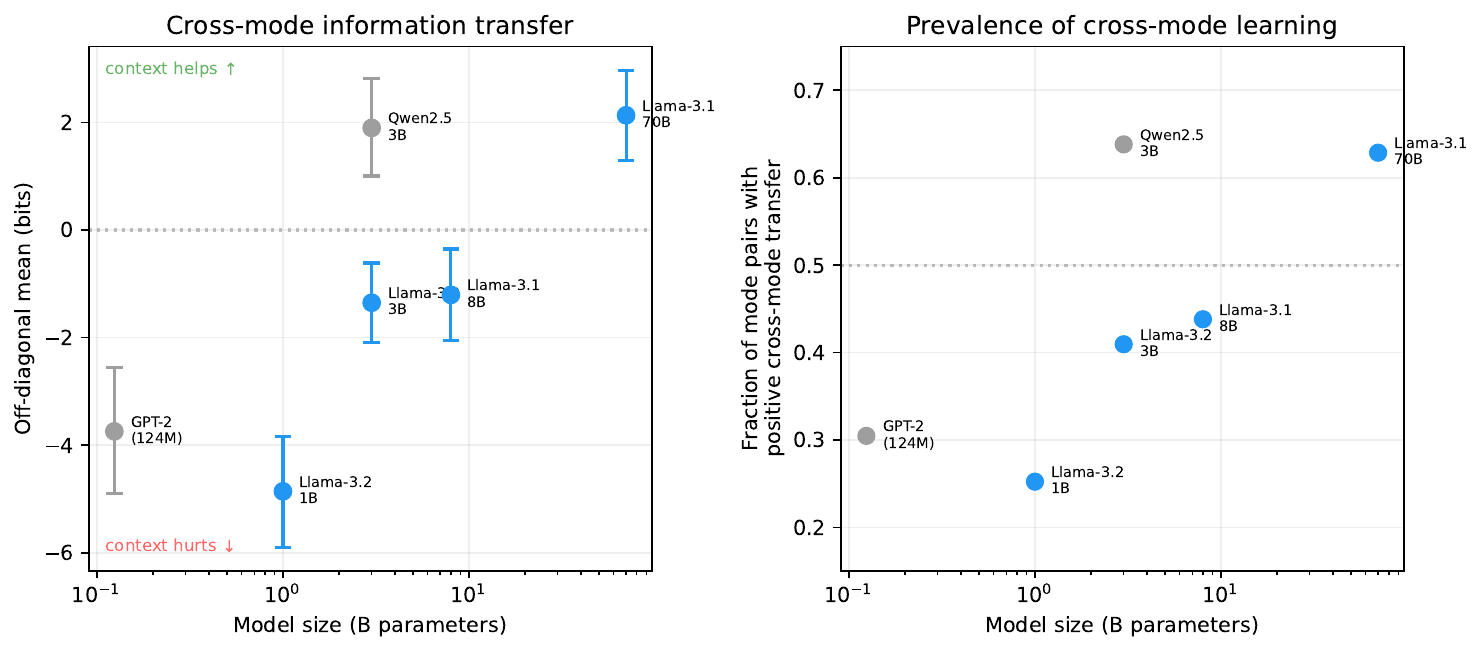}
\caption{Cross-mode information transfer scales with model size. \textbf{Left}: Mean off-diagonal surprise reduction (with bootstrap 95\% CI) transitions from negative to positive as models grow, indicating that larger models extract more information from cross-mode context. \textbf{Right}: The fraction of mode pairs showing positive cross-mode transfer increases monotonically across the \scalingNumLlamaModels{} Llama models (blue; $p = \scalingFactorialProbPct\%$ under monotone null).
GPT-2 and Qwen2.5-3B (gray) are included for context but differ in architecture.}
\label{fig:scaling-crossmode}
\end{figure*}

\paragraph{Cross-mode information scales with model quality.} We test how cross-mode information transfer varies with $\theta$'s scale by running the pairwise matrix experiment on \scalingNumLlamaModels{} dense Llama models (1B, 3B, 8B, 70B) using the same \crossmodeNumModes{} modes and 5 samples per mode.
Figure~\ref{fig:scaling-crossmode} shows the results.
The off-diagonal mean increases monotonically with model size: $\scalingOffDiagLlamaOneB$ (1B), $\scalingOffDiagLlamaThreeB$ (3B), $\scalingOffDiagLlamaEightB$ (8B), $\scalingOffDiagLlamaSeventyB$ (70B) bits, transitioning from negative (cross-mode context hurts) to positive (cross-mode context helps).
The fraction of positive off-diagonal entries follows the same monotone trend: \scalingFracPosLlamaOneB\%, \scalingFracPosLlamaThreeB\%, \scalingFracPosLlamaEightB\%, \scalingFracPosLlamaSeventyB\%.
This hypothesis was pre-registered before running any Llama models, and the probability of accidental monotonicity across \scalingNumLlamaModels{} models under the null is $1/\scalingNumLlamaModels! \approx \scalingFactorialProbPct\%$.

\paragraph{The framework admits other metrics.} The $a_k$ curve is the primary object of our framework, and $D_{Ca_n} = C \times a_n$ is one natural summary of it among many.
Reporting the curve directly preserves information that any single scalar (including ours) collapses.
Alternative scalars could be derived from the same curve to capture different desiderata: weighted sums of $a_k - a_\infty$ (emphasizing early or late positions), the slope or curvature at a specific $k$ (measuring how quickly $\theta$ learns), curve-shape descriptors, or coherence-weighted variants that combine $C$ with quantities other than $a_n$.
Different applications may warrant different choices; for instance, a policy-comparison setting might prefer a metric sensitive to \emph{changes} in the curve shape, while a ranking setting needs a single well-behaved scalar.
We have not explored this space systematically; we focus on $D_{Ca_n}$ because it empirically works well across regimes and has a clean information-theoretic interpretation, but we expect the framework to support a family of related metrics with complementary properties.
One untested family is coherence-reweighting: a practitioner with a clear preference for sharper Mixed penalisation could replace $C \times a_n$ with $C^{\alpha} \times a_n$ for $\alpha > 1$ to drive low-coherence sets toward zero faster.
We have not evaluated such variants on the Tevet or OLMo experiments and so make no empirical claim about them.

\paragraph{Bits/byte vs.\ total bits, and the length-matching consequence.}
A separate choice within the framework is whether to report total bits (length-dependent but tokenizer-agnostic) or per-byte bits/byte (also tokenizer-agnostic, with length scaled out).
We default to bits/byte for the headline diversity score, but per-byte quantities are not invariant to response length: in a causal LM, longer responses give $\theta$ more within-response context to predict later tokens, lowering per-byte surprise.
For the OLMo-2-7B experiment (Section~\ref{sec:rlhf-experiment}) raw response length is not monotone across stages, so we score the length-matched subset as the headline analysis: every response in a (stage, prompt) tuple is truncated to a common per-prompt UTF-8 byte length (the minimum across all 40 responses) before scoring.
$111/300$ prompts had a per-prompt common length below 50 bytes -- driven by the base model on AlpacaEval (no instruction-following prior) and by SFT on NB-curated short-answer prompts -- and were dropped because length-matching them would compare essentially-empty fragments.
The headline numbers therefore condition on the $150/200$ AlpacaEval and $39/100$ NB-curated prompts where length-matching is well-defined; the dropped prompts are themselves a stage-specific behavioural signal we are not measuring here.
We also verified that the monotone $D$ drop holds on the un-truncated $200+100$ prompt sets, with all three pre-registered H1 contrasts remaining Bonferroni-significant at $p<10^{-13}$ on both prompt sets.
Full numbers and reproduction commands are in \path{investigations/length_matched_rlhf/VERDICT.md}.

%% file: sections/09_future_work.tex
\section{Future Work}

Several directions remain unexplored.

\paragraph{Prompt format optimization.} We use a fixed prompt template throughout and do not attempt to optimize it for ICL elicitation.
Prompt engineering could improve the metric's discriminative power.
Dimension-specific framings (e.g., ``Here are several stories in different genres'') are one such variant: priming $\theta$ to expect variation along a dimension removes the small portion of the surprise attributable to the mere existence of that variation, but it does not isolate per-dimension diversity, since the rest of the cross-response surprise reduction remains.

\paragraph{Ensembling base models.} A single base model $\theta$ may produce non-monotone $a_k$ curves when pushed out of distribution by long contexts.
Ensembling multiple base models at the token level (averaging softmax probabilities) could stabilize the estimates while preserving the autoregressive structure.
Our codebase supports token-level ensembling, but we have not yet validated it experimentally.

\paragraph{Disagreement cases with other metrics.}
We report aggregate correlations with embedding-based and reference metrics on Tevet's tasks (Section~\ref{sec:tevet}) but do not look at \emph{which} response sets the metrics rank differently.
Identifying prompts or response sets where $C \times a_n$ disagrees with SentBERT, distinct-$n$, or McDiv is the most direct way to characterise what each metric is actually measuring; the disagreements are where the choice of metric matters.
We expect two regimes to drive disagreement: response sets that share arbitrary rare patterns a base model can pick up on but that fall outside what a fixed sentence embedder represents (where $C \times a_n$ should detect repetition that embeddings miss), and response sets where one response is a mixture of two others (where progressive conditioning credits the mixture for being predictable from its components, while pairwise embedding-distance averages do not).

\paragraph{Total-bits variants of $D$.}
We report $D_{Ca_n} = C \times a_n$ in bits/byte for length control, normalising by bytes rather than tokens so that scores remain tokenizer-agnostic.
The implementation also exposes $a_n$ in total bits, which is likewise tokenizer-agnostic but not length-controlled.
A hybrid scalar $C \times a_n^{\text{bits}}$ rewards length (longer responses have more positions at which to be surprising), which some practitioners may want; characterising when each variant is the appropriate measurement target is left to future work.

\paragraph{Optimizing generators against $D$.}
We use $C \times a_n$ as an evaluation-time diagnostic, but it could in principle serve as a training or decoding objective. Future work could fine-tune or sample from a generator $\pi$ with the goal of maximizing $C \times a_n$ under a fixed base model $\theta$ to push $\pi$ toward response sets that are both coherent and diverse.

%% file: sections/appB_excess_entropy.tex
\section{Excess Entropy and the \texorpdfstring{$C \times E$}{C x E} Score}\label{app:excess-entropy}

This appendix develops the excess entropy $E$, an alternative summary of the $a_k$ curve that is theoretically interesting (connecting to the excess entropy of computational mechanics \citep{crutchfield2003regularities} and to the total correlation) but empirically inferior to $D_{Ca_n}$ on the external benchmarks of Section~\ref{sec:tevet}.
We tried both the last-point estimator $\hat E_n$ and a sigmoid-extrapolated projected floor for $a_\infty$ (Section~\ref{app:excess-entropy-def}); we stopped using $E$ when it became clear it cannot distinguish diverse from non-diverse response sets in the few-draws regime where modes do not repeat within $n$ samples.
We include it here for theoretical completeness, as a null result, and to explain the empirical comparisons that motivate our preference for $D_{Ca_n}$.

\subsection{The Excess Entropy}\label{app:excess-entropy-def}

The primary metric $D_{Ca_n} = C \times a_n$ uses the curve's endpoint $a_n$: as the surprise after conditioning on $n-1$ peer responses, $a_n$ reflects both the curve's level and the drop from $a_1$ in a single scalar.
Section~\ref{sec:tevet} shows this combination is what discriminates diverse from non-diverse response sets in the few-draws regime, while $a_1$ alone (the level) and $E$ (the integrated drop, defined below) each underperform.

We thought that $E$'s learnable structure would be interesting and track diversity.
Following the concept of excess entropy from computational mechanics \citep{crutchfield2003regularities}, define the \textbf{excess entropy}:
\begin{equation}\label{eq:excess}
    E = \sum_{k=1}^{\infty}\bigl(a_k - a_\infty\bigr)
\end{equation}

Units: bits.
This is the total learnable structural information in the progressive conditioning process, above the irreducible residual noise.
It converges whenever $\theta$'s surprise at new responses eventually stabilizes (since $a_k \to a_\infty$ and the excess $e_k = a_k - a_\infty$ decays to zero).
The ``structure'' captured by $E$ is not limited to discrete mode identity: it includes any inter-response regularity to which $\theta$ assigns lower conditional surprise.

In practice, $a_\infty$ is unknown.
We estimate it as $a_n$ (the last observed value) and compute:
\begin{equation}\label{eq:excess-hat}
    \hat{E}_n = \sum_{k=1}^{n}(a_k - a_n)
\end{equation}

This underestimates $E$ (since $a_n \geq a_\infty$), but the bias decreases with $n$.

\paragraph{Parametric extrapolation of $a_\infty$.} Rather than using $a_n$ directly, one can fit a parametric model to the observed curve and extrapolate $a_\infty$.
The $a_k$ curve is theoretically expected to be sigmoidal (with concave-up/exponential decay as the degenerate case when the inflection point $k_0 < 1$).
This motivates fitting a four-parameter sigmoid:
\begin{equation}\label{eq:sigmoid-fit}
    a_k = a_\infty + \frac{\alpha}{1 + e^{\beta(k - k_0)}}
\end{equation}
where $a_\infty$ is the asymptotic floor, $\alpha$ is the total drop from $a_1$ to $a_\infty$, $\beta > 0$ controls the steepness of the transition, and $k_0$ is the inflection point.
The fit yields $a_\infty$ without requiring $n$ to be large enough for convergence.
Section~\ref{sec:mode-count-scaling} provides empirical evidence that the sigmoid-extrapolated $E_{\mathrm{fit}}$ recovers the expected monotonic relationship between mode count and excess entropy, while the raw $\hat{E}_n$ estimator does not.

\paragraph{Relationship to total correlation.} The excess entropy and total correlation are complementary decompositions of the same underlying structure.
Working in expectation over i.i.d.\ draws from $\pi$, let $\bar{a}_1 = \E[-\log_2 \theta(r \mid p)]$ denote the expected unconditional cross-entropy, $\bar{a}_k = \E[a_k]$ the expected conditional cross-entropy at position $k$, and $e_k = \bar{a}_k - a_\infty$ the excess at step $k$.
The per-step mutual information is $I_k = \bar{a}_1 - \bar{a}_k = (\bar{a}_1 - a_\infty) - e_k$.
Summing:
\begin{equation}\label{eq:tc-decomp}
    \TC_n = \sum_{k=1}^{n} I_k = n(\bar{a}_1 - a_\infty) - E_n
\end{equation}

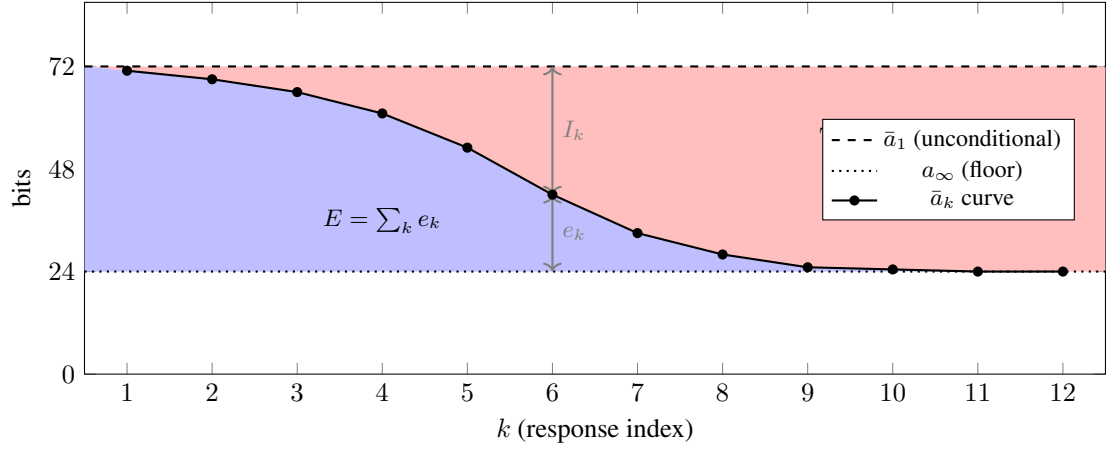
\begin{figure*}[tb]
\centering
\begin{tikzpicture}
\begin{axis}[
    width=0.88\textwidth,
    height=6.5cm,
    xlabel={$k$ (response index)},
    ylabel={bits},
    xmin=0.5, xmax=12.5,
    ymin=0, ymax=87,
    xtick={1,2,3,4,5,6,7,8,9,10,11,12},
    ytick={0, 24, 48, 72},
    legend style={font=\small, at={(0.97,0.55)}, anchor=east},
    clip=false,
]

% Define h-bar line
\addplot[name path=hbar, forget plot, draw=none]
    coordinates {(0.5,72) (12.5,72)};

% Define a_infinity line
\addplot[name path=ainf, forget plot, draw=none]
    coordinates {(0.5,24) (12.5,24)};

% Define a_k curve (theoretical, in expectation)
\addplot[name path=ak, forget plot, draw=none]
    coordinates {
        (0.5,72) (1,71) (2,69) (3,66) (4,61)
        (5,53) (6,42) (7,33) (8,28) (9,25)
        (10,24.5) (11,24) (12,24) (12.5,24)
    };

% Fill TC area (between h-bar and a_k) — red
\addplot[red!25, forget plot] fill between[of=hbar and ak];

% Fill E area (between a_k and a_infinity) — blue
\addplot[blue!25, forget plot] fill between[of=ak and ainf];

% Draw the visible lines on top
\addplot[thick, dashed, black]
    coordinates {(0.5,72) (12.5,72)};
\addlegendentry{$\bar{a}_1$ (unconditional)}

\addplot[thick, dotted, black]
    coordinates {(0.5,24) (12.5,24)};
\addlegendentry{$a_\infty$ (floor)}

\addplot[thick, black, mark=*, mark size=1.5pt]
    coordinates {
        (1,71) (2,69) (3,66) (4,61)
        (5,53) (6,42) (7,33) (8,28) (9,25)
        (10,24.5) (11,24) (12,24)
    };
\addlegendentry{$\bar{a}_k$ curve}

% Labels for the colored regions
\node[font=\small] at (axis cs:10,55) {$\TC_n = \sum_k I_k$};
\node[font=\small] at (axis cs:4,36) {$E = \sum_k e_k$};

% Annotation arrows showing I_k and e_k for a single step (k=6, a_6=42)
\draw[<->, thick, gray] (axis cs:6,42) -- node[right, font=\footnotesize] {$e_k$} (axis cs:6,24);
\draw[<->, thick, gray] (axis cs:6,72) -- node[right, font=\footnotesize] {$I_k$} (axis cs:6,42);

\end{axis}
\end{tikzpicture}
\caption{Decomposition of the gap between unconditional surprise $\bar{a}_1$ and the asymptotic floor $a_\infty$ at each step $k$.
The per-step gap $\bar{a}_1 - a_\infty$ splits into mutual information $I_k = \bar{a}_1 - \bar{a}_k$ (\textcolor{red!70}{red}, above the curve) and excess $e_k = \bar{a}_k - a_\infty$ (\textcolor{blue!70}{blue}, below the curve).
Summing across $k$: the red area is the total correlation $\TC_n$; the blue area is the excess entropy $E$.
As $k$ grows, $e_k \to 0$ and each step contributes the full $\bar{a}_1 - a_\infty$ to $\TC_n$, so $\TC_n$ grows linearly while $E$ converges.
All quantities are in bits.
This figure is theoretical: the sigmoidal $\bar{a}_k$ shape reflects our initial expectation, but we subsequently found that the empirical curves are exponential decay without an initial plateau (see Section~\ref{sec:cross-mode-learning}).}
\label{fig:decomposition}
\end{figure*}

For large $n$ where $E_n \to E$, the total correlation grows linearly with slope $(\bar{a}_1 - a_\infty)$: $\TC_n \approx n(\bar{a}_1 - a_\infty) - E$.
Figure~\ref{fig:decomposition} illustrates.
The three quantities have clean interpretations, all in bits: $(\bar{a}_1 - a_\infty)$ is the per-response redundancy once $\theta$ has fully learned the available structure; $E$ is the total finite structural information, consumed during $\theta$'s transient learning phase; $\TC_n$ is the cumulative redundancy, which grows without bound.

\subsection{Per-Byte Excess Entropy Rate}\label{app:per-byte-rate}

The excess entropy $E = \sum_k (a_k - a_\infty)$ is in bits.
For a length-normalized variant, we define the \textbf{per-byte excess entropy rate} by normalizing each response's surprise by its byte count \emph{before} averaging across permutations.
For a single permutation $\sigma$, the per-byte conditional surprise at position $k$ is $a_k^\sigma / \|r_{\sigma(k)}\|$ (bits/byte).
Averaging over permutations gives
\begin{equation}
    \hat{a}_k^{\mathrm{rate}} = \E_\sigma\!\left[\frac{a_k^\sigma}{\|r_{\sigma(k)}\|}\right]
\end{equation}
and the per-byte floor is estimated analogously from the last position: $\hat{a}_\infty^{\mathrm{rate}} = \hat{a}_n^{\mathrm{rate}}$.
The per-byte excess entropy rate is
\begin{equation}\label{eq:per-byte-E}
    E_{\mathrm{rate}} = \sum_{k=1}^{n}\bigl(\hat{a}_k^{\mathrm{rate}} - \hat{a}_\infty^{\mathrm{rate}}\bigr).
\end{equation}
This treats each response equally regardless of length.
Note that $E_{\mathrm{rate}} \neq E / \bar{B}$ in general: because total surprise $a_k$ and byte count $\|r_k\|$ are positively correlated, $E / \bar{B}$ overweights long responses, while the per-response normalization in $E_{\mathrm{rate}}$ avoids this.

\subsection{The \texorpdfstring{$C \times E$}{C x E} Score}

Combining $E$ with the coherence term $C$ (Section~\ref{sec:coherence}) yields scalar scores
\begin{equation}\label{eq:D-total}
    C \times E \quad \text{(bits)} \qquad \text{or} \qquad C \times E_{\mathrm{rate}} \quad \text{(bits/byte)}.
\end{equation}
The first measures total structural information, and the second normalizes per byte so long and short responses are treated equally.

\subsection{Why Not Weight Excess Entropy Inside the Sum?}\label{app:why-not-weight}

A natural alternative to reporting $E$ and $C$ separately is to weight the terms within $E$ by coherence, computing $E_w = \sum_k w_k \cdot (a_k - a_\infty)$ where $w_k = 2^{-h_\theta(r_k \mid p)}$.
This would suppress contributions from incoherent modes directly at the point of measurement.
We considered and rejected this approach because it breaks the information-theoretic interpretation. $E = \sum_k e_k$ has a clean meaning as the total structural information above the noise floor, connected to Crutchfield and Feldman's excess entropy, and inserting weights makes $E_w$ a hybrid quantity that is not the excess entropy of any well-defined process.

\subsection{Limitations of $C \times E$}

$C \times E$ inherits two limitations from $E$.
First, $E$ measures \emph{recurrence}: surprise reduction as multiple responses accumulate, rather than continuous spread.
A policy that draws from a broad, roughly continuous distribution of coherent outputs has $a_k \approx a_1$ for all $k$, yielding $E \approx 0$ despite maximal diversity.
Second, $E$ is near-zero in the few-draws regime: with only 5--10 responses and no repeated modes, $\theta$ finds little learnable structure regardless of diversity, so $C \times E$ carries almost no signal.
These limitations are the reason $D_{Ca_n}$ is preferred as the primary score: $a_\infty$ is the \emph{level} of the floor, which differs meaningfully between diverse and non-diverse response sets even when the curve stays nearly flat.

\begin{figure*}[!t]
\centering
\includegraphics[width=\textwidth]{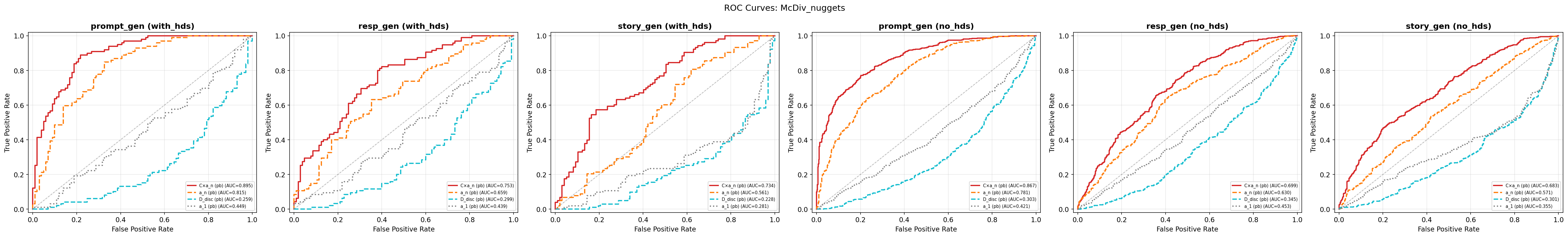}
\caption{ROC curves for McDiv\_nuggets binary classification (Qwen2.5-3B, 50 permutations).
Five per-byte metrics are overlaid per panel: $C \times a_n$ (red, solid), $a_n$ alone (orange, dashed), $D_{\mathrm{fit}} = C \times E_{\mathrm{fit}}$ (blue, solid), $D_{\mathrm{disc}} = C \times \hat{E}_n$ (cyan, dashed), and $a_1$ (gray, dotted). Line style is redundant with color so the encoding remains legible under common color-vision deficiencies.
$C \times a_n$ dominates the other four scores on every task; the $E$-based scores ($D_{\mathrm{fit}}, D_{\mathrm{disc}}$) hug or fall below the chance diagonal in this 5-response regime.}
\label{fig:roc-curves}
\end{figure*}

\paragraph{Empirical inferiority.} On Tevet's diversity-eval benchmark with only 5 responses per set (Section~\ref{sec:tevet}), $C \times a_n$ achieves ROC AUC of \tevetMcDivNugPromptGenCxAnAUC{} on McDiv\_nuggets prompt\_gen while the $D_{\mathrm{fit}} = C \times E_{\mathrm{fit}}$ variant cannot be computed at all (the four-parameter sigmoid fails to converge on only five points) and the discretized $D_{\mathrm{disc}}$ variant is anti-correlated with diversity (AUC = \tevetMcDivNugPromptGenDdiscAUC{}, well below chance).
Figure~\ref{fig:roc-curves} shows the ROC curves.

\paragraph{Why $E$ fails in this regime.} With only 5 responses and no repeated modes, $\theta$ finds little learnable structure regardless of diversity.
The $a_k$ curve stays approximately flat, yielding $E \approx 0$ for both diverse and non-diverse sets.
The $E$-based scores thus carry almost no signal.
The asymptotic floor $a_\infty$, by contrast, differs substantially between the two conditions: diverse responses remain surprising after conditioning ($a_\infty$ high), while paraphrases become predictable ($a_\infty$ low).
This is the fundamental reason $D_{Ca_n}$ is preferred.

% D.6 "Coherence Normalization Strategies" excluded from rendered paper:
% strategies were considered but never used in our experiments. Source
% preserved here; flip \iffalse to \iftrue to re-include. The original
% (non-workshop) paper references this via \ref{app:c-normalization} in
% 05_reporting.tex; that reference will dangle if the original paper is
% rebuilt without re-including this block.
\iffalse
\subsection{Coherence Normalization Strategies}\label{app:c-normalization}

Two strategies are available when coherence variation dominates diversity variation:
\begin{enumerate}[nosep]
    \item \textbf{Normalization}: Replace $C$ with $C / C_{\mathrm{base}}$, where $C_{\mathrm{base}}$ is the coherence of $\theta$'s own samples given the same prompt.
The ratio lives in $(0, 1]$ and equals 1 when $\pi$ is as coherent as $\theta$, removing the baseline suppression.
This has a clean interpretation as ``what fraction of baseline coherence does $\pi$ retain?''
    \item \textbf{Fractional exponent}: Replace $C$ with $C^\alpha$ for $\alpha \in (0, 1)$, which compresses the coherence penalty.
This softens the dominance of $C$ but introduces a free hyperparameter without information-theoretic motivation.
\end{enumerate}
We recommend normalization when a single-scalar ranking is the primary use case, and the unmodified $C$ when the goal is diagnostic.
We did not use either strategy in our experiments.
\fi

\subsection{Mode Count Scaling: Excess-Entropy Metrics}

\begin{figure}[!htbp]
\centering
\includegraphics[width=\columnwidth]{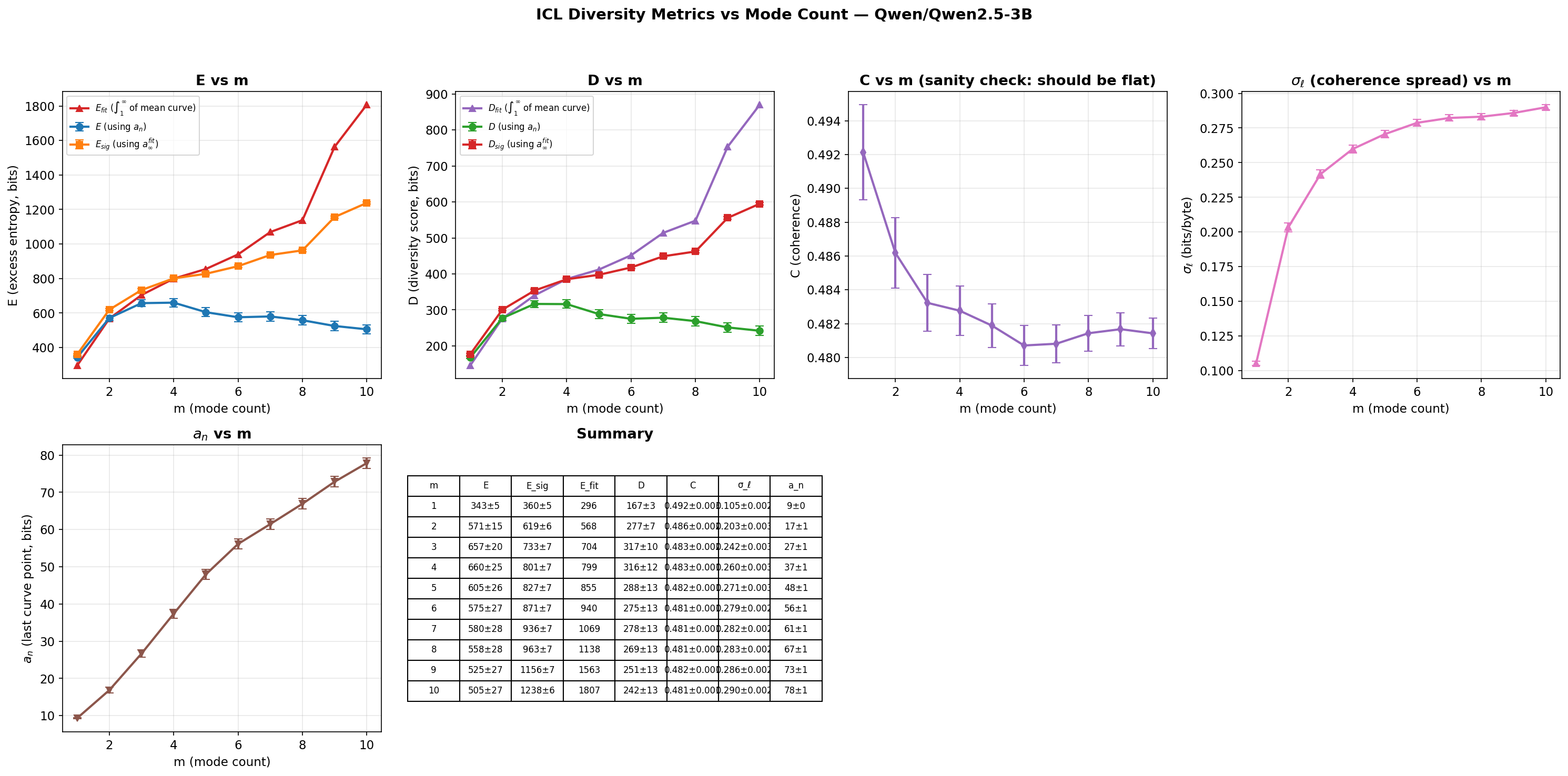}
\caption{Summary metrics vs.\ mode count $m$ (Qwen2.5-3B, $n=\modeCountNResponses$, \modeCountNDraws{} draws). $E_{\mathrm{fit}}$ (sigmoid-extrapolated) increases monotonically, while the raw $\hat{E}_n$ does not.}
\label{fig:metrics-vs-m}
\end{figure}

In the synthetic mode count experiments (Section~\ref{sec:mode-count-scaling}), the sigmoid-extrapolated $E_{\mathrm{fit}}$ is monotonic in mode count for Qwen2.5-3B but has wide confidence intervals at high $m$, while the raw $\hat{E}_n$ estimator is non-monotonic.
Table~\ref{tab:mode-count} and Figure~\ref{fig:metrics-vs-m} show the details.

\begin{table}[!htbp]
\centering
\caption{Mode count scaling metrics for Qwen2.5-3B ($n=\modeCountNResponses$, \modeCountNDraws{} draws). $E_{\mathrm{fit}}$ is the sigmoid-extrapolated excess entropy; $a_n$ is the mean floor at $k=\modeCountNResponses$; $\sigma_\ell$ is per-byte coherence spread; $k_0$ is the sigmoid inflection point.
Confidence intervals are 95\% bootstrap. The high-$m$ rows ($m{\ge}8$) have $E_{\mathrm{fit}}$ CIs spanning more than half the mean and should be read as preliminary; the trend direction is robust but the point values are imprecise.}
\label{tab:mode-count}
\small
\input{results/tables/mode_count_scaling.tex}
\end{table}

\begin{figure}[!htbp]
\centering
\includegraphics[width=\columnwidth]{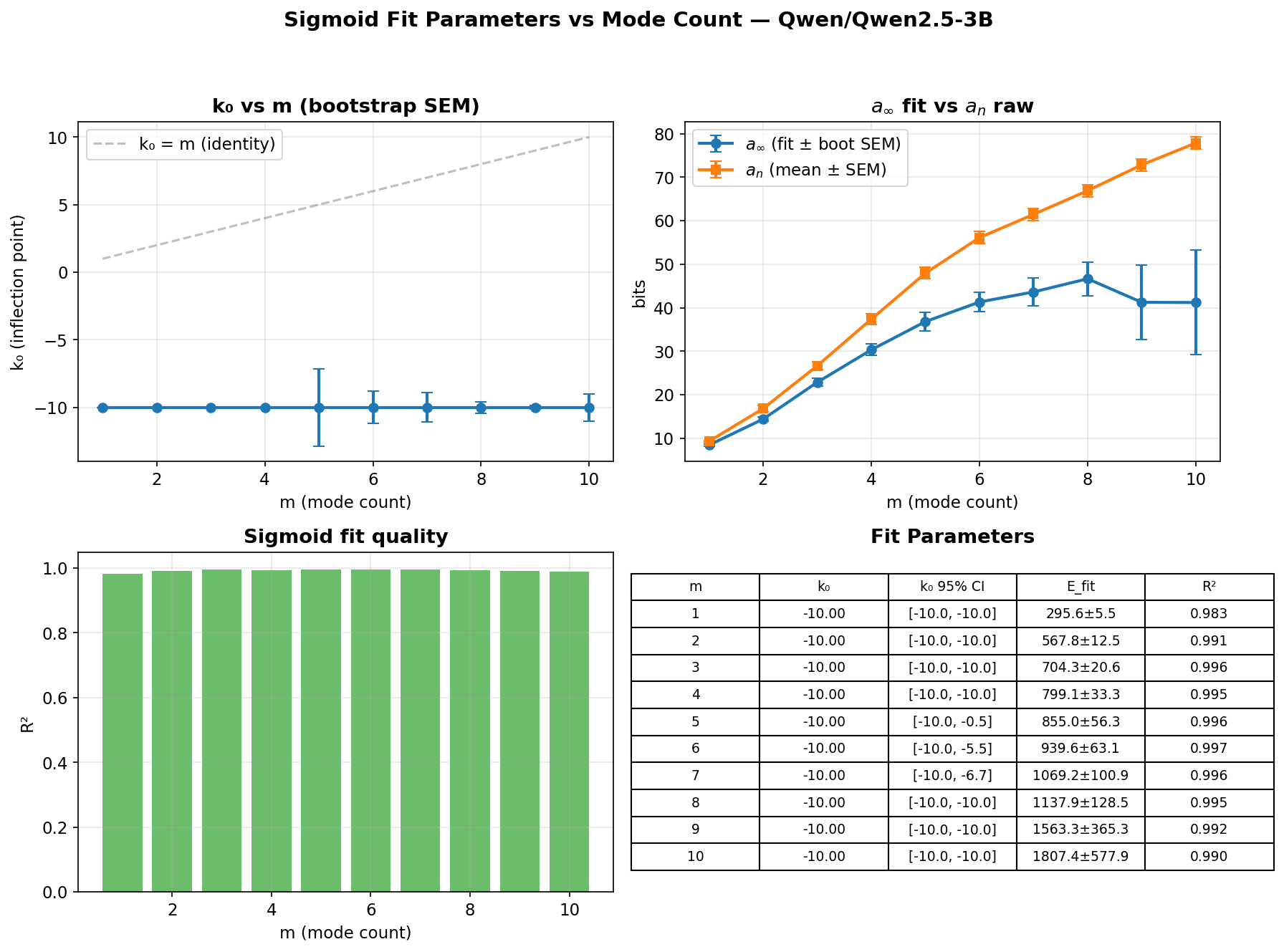}
\caption{Sigmoid fit parameters vs.\ mode count $m$ (Qwen2.5-3B).
The inflection point $k_0$ remains at the lower bound ($-10$) across all $m$, indicating pure exponential decay without an initial plateau (see Section~\ref{sec:cross-mode-learning}).}
\label{fig:fit-params}
\end{figure}

\input{results/tables/confound_examples.tex}

%% file: results/tables/mode_count_scaling.tex
% Generated by: scripts/generate_mode_count_table.py
% Data source: results/mode_count/qwen2.5-3b_1k_draws.json
%              results/mode_count/qwen2.5-3b_fits.json
\begin{tabular}{@{}r rrrr@{}}
\toprule
$m$ & $E_{\mathrm{fit}}$ (bits) & $a_n$ (bits) & $\sigma_\ell$ (bits/byte) & $k_0$ \\
\midrule
1  & $296 \pm 11$ & 9.3 & 0.1 & $-10.0$ \\
2  & $568 \pm 24$ & 16.9 & 0.2 & $-10.0$ \\
3  & $704 \pm 40$ & 26.7 & 0.2 & $-10.0$ \\
4  & $799 \pm 67$ & 37.4 & 0.3 & $-10.0$ \\
5  & $855 \pm 113$ & 48.0 & 0.3 & $-10.0$ \\
6  & $940 \pm 127$ & 56.1 & 0.3 & $-10.0$ \\
7  & $1069 \pm 194$ & 61.5 & 0.3 & $-10.0$ \\
8  & $1138 \pm 253$ & 66.9 & 0.3 & $-10.0$ \\
9  & $1563 \pm 705$ & 72.8 & 0.3 & $-10.0$ \\
10  & $1807 \pm 1194$ & 77.8 & 0.3 & $-10.0$ \\
\bottomrule
\end{tabular}

%% file: results/tables/confound_examples.tex
% Generated by scripts/dataset_confound_analysis.py
% Per-byte a_1 for r_0 of the illustrative high- and low-diversity samples.
% high sample: test.sa-sb.sa::00698
% low sample:  test.sa-sb.sa-b::00279
\newcommand{\confoundHighRoAone}{0.783}
\newcommand{\confoundLowRoAone}{1.777}
\newcommand{\confoundHighSampleId}{test.sa-sb.sa::00698}
\newcommand{\confoundLowSampleId}{test.sa-sb.sa-b::00279}

%% file: sections/appC_mcdiv_confounds.tex
\section{Dataset Construction Confounds in McDiv}\label{app:confound}

\begin{figure*}[!t]
\centering
\includegraphics[width=\textwidth]{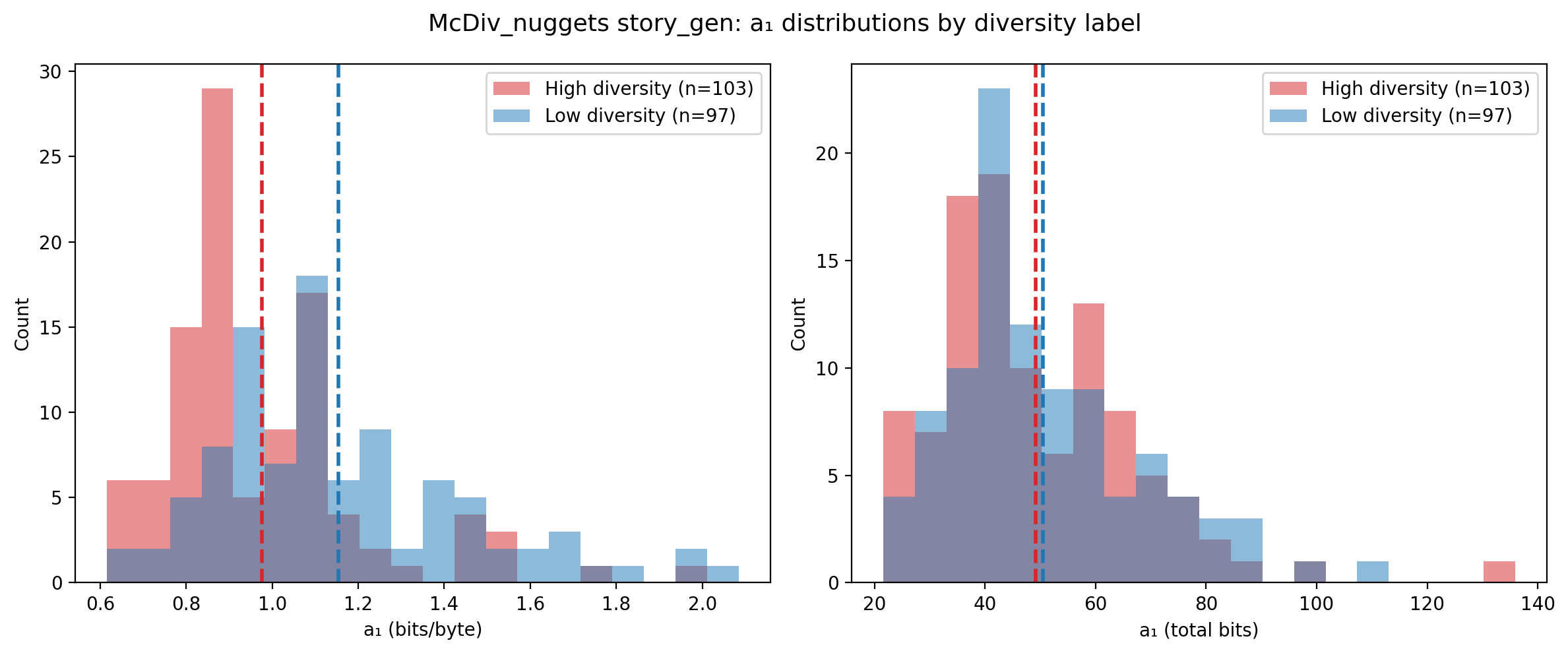}
\caption{Distribution of $a_1$ (unconditional surprise of the first response) for high- vs.\ low-diversity story\_gen samples.
Left: per-byte.
Right: total bits.
The per-byte gap (see Table~\ref{tab:confound-stats}) is a dataset construction artifact.}\label{fig:confound-a1-dist}
\end{figure*}

When validating against the McDiv\_nuggets benchmark \citep{tevet2020evaluating}, we observed that the mean $\bar{a}_1$ curve (per-byte) for the ``low diversity'' group starts \emph{above} the ``high diversity'' group in the story\_gen task (Figure~\ref{fig:confound-a1-dist}).
Since $a_1$ measures the surprise of the first response conditioned only on the prompt (before any other responses appear in context), this gap cannot reflect diversity.
It is a confound in the dataset construction.

\subsection{Mechanism}

The McDiv protocol (Tevet and Berant \S6.4), from which McDiv\_nuggets is sampled (Tevet and Berant Appendix C.2 specifies McDiv\_nuggets as the 3K subset of McDiv on which distinct-$n$ correlates at zero), pairs a low-diversity set with each high-diversity set as follows.
Workers first write five \emph{different} continuations (the high-diversity set).
The same worker is then asked to \emph{self-select one} of their own five responses and paraphrase it five times, preserving content while varying form (the low-diversity set).
Tevet and Berant do not characterize the distribution of which responses workers self-select.

Empirically (see Table~\ref{tab:confound-stats} and Figure~\ref{fig:confound-a1-dist}, and the illustrative examples from the McDiv\_nuggets story\_gen dataset in Section~\ref{sec:mcdiv-examples} below), we observe that the self-selected endings for the low-diversity sets tend to be more specific or dramatic (e.g., ``He scored winner'' or ``they quit'') than the typical high-diversity continuations (e.g., ``Joel fired the cook when things went too far downhill'').
This means that individual low-diversity responses are intrinsically more surprising to the base model, not because of diversity, but because of the specificity of the endings workers self-selected for paraphrasing.

\subsection{Evidence}

Table~\ref{tab:confound-stats} summarizes the gap.
The per-byte $a_1$ difference is substantial (a content effect) while the total-bits difference is small because high-diversity responses are on average several bytes longer.

\begin{table}[!htbp]
\centering
\caption{Unconditional surprise ($a_1$) by diversity label, McDiv\_nuggets story\_gen.}\label{tab:confound-stats}
\input{results/tables/confound_stats.tex}
\end{table}

The gap persists after binning by response length (Table~\ref{tab:confound-length}), confirming it is primarily a content effect rather than a length artifact: the high-minus-low per-byte gap remains positive across the bulk of the length range.

\begin{table}[!htbp]
\centering
\caption{Unconditional per-byte surprise by response length bin, story\_gen.}\label{tab:confound-length}
\input{results/tables/confound_length.tex}
\end{table}

\subsection{Illustrative Examples}\label{sec:mcdiv-examples}

\begin{figure*}[!t]
\centering
\includegraphics[width=\textwidth]{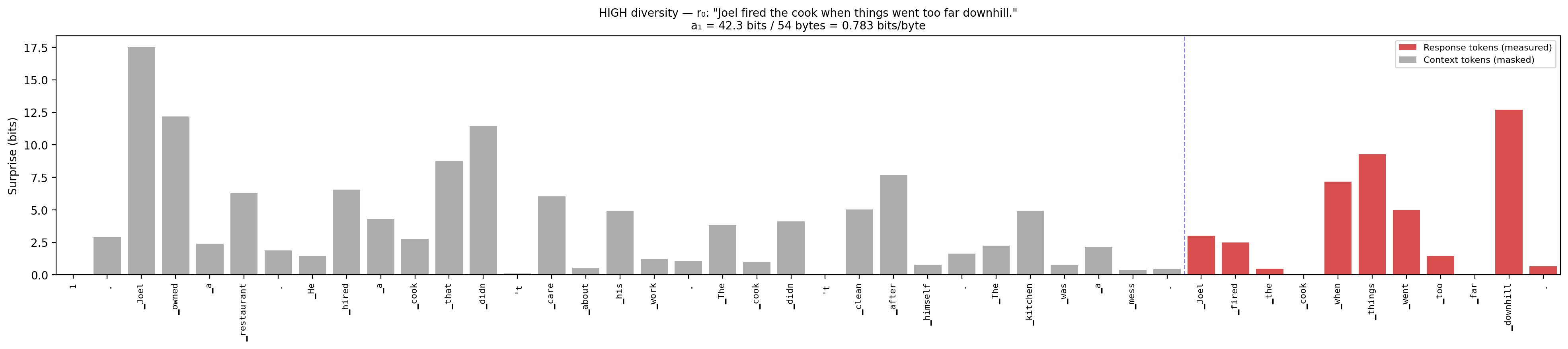}
\caption{Per-token surprise for a high-diversity sample's first response.
The continuation (``Joel fired the cook...'') is predictable, yielding low per-token surprise across response tokens.}\label{fig:confound-high}
\end{figure*}

\begin{figure*}[!t]
\centering
\includegraphics[width=\textwidth]{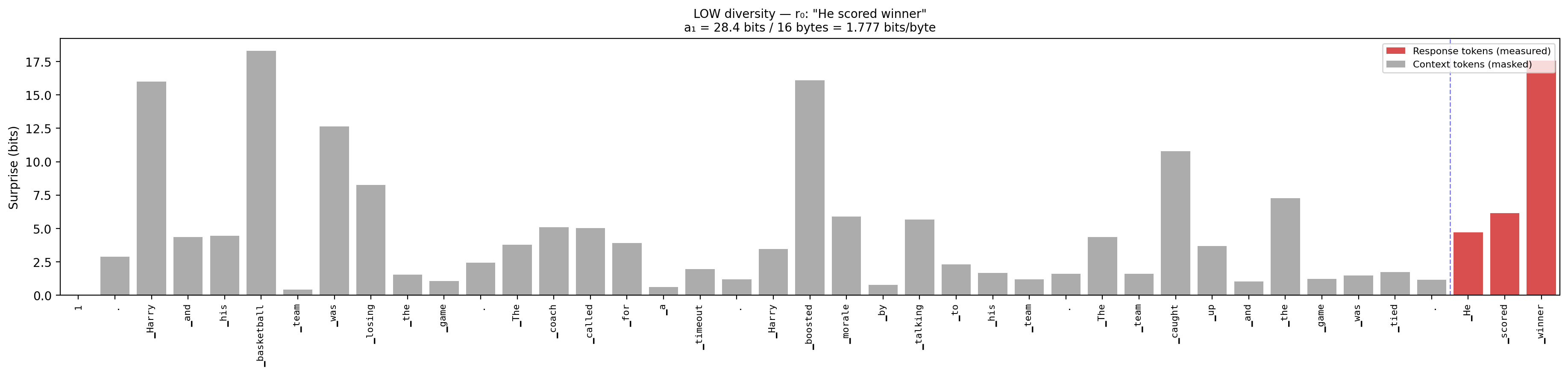}
\caption{Per-token surprise for a low-diversity sample's first response.
The specific ending (``He scored winner'') is inherently more surprising to the base model, despite being labeled ``low diversity.''}\label{fig:confound-low}
\end{figure*}

Figures~\ref{fig:confound-high} and~\ref{fig:confound-low} show per-token surprise (in bits) for the first response of a high-diversity and low-diversity sample, respectively.
Samples are drawn by ranking each label group by per-byte $a_1$ (ascending for high-diversity, descending for low-diversity) and taking the sample at the $\lfloor N/4 \rfloor$ position, chosen to be representative of the expected confound direction rather than an extreme outlier; the specific pinned IDs (\texttt{sa\_00698}, \texttt{sa-b\_00279}) are fixed in \texttt{scripts/dataset\_confound\_analysis.py} for reproducibility.
Red bars are measured response tokens; grey bars are masked context tokens.

\paragraph{High diversity} (Figure~\ref{fig:confound-high}).
Context: \emph{``Joel owned a restaurant.
He hired a cook that didn't care about his work.
The cook didn't clean after himself.
The kitchen was a mess.''} Response: \emph{``Joel fired the cook when things went too far downhill.''} This is a natural, predictable continuation ($a_1 = \confoundHighRoAone$ bits/byte).

\paragraph{Low diversity} (Figure~\ref{fig:confound-low}).
Context: \emph{``Harry and his basketball team was losing the game.
The coach called for a timeout.
Harry boosted morale by talking to his team.
The team caught up and the game was tied.''} Response: \emph{``He scored winner.''} This is a specific, somewhat ungrammatical ending ($a_1 = \confoundLowRoAone$ bits/byte).
All five responses in this set are paraphrases of the same idea (``Harry scored the winning point'').

\subsection{Implications}

The construction confound (low-diversity sets are paraphrases of specific dramatic endings rather than a random subsample of low-diversity content) lifts the entire $\bar{a}_k$ curve on low-diversity sets, since their content is intrinsically surprising under $\theta$ at every $k$, and pushes $C$ down because the same content is less coherent.
$a_1$'s upward shift opposes the diversity label; $C$'s downward shift happens to align with it (a confound artifact, not coherence genuinely tracking diversity).
At $a_n$, the diversity signal we are measuring overrides the confound's upward pull: paraphrases collapse to predictable once $\theta$ has seen the others while genuinely diverse responses do not, so $a_n$ ends up lower on low-diversity sets and $C \times a_n$ tracks the label despite the confound (Section~\ref{sec:tevet}).

%% file: results/tables/confound_stats.tex
% Generated by scripts/dataset_confound_analysis.py
% Source: results/tevet/qwen25_completion_v3/McDiv_nuggets/*_with_hds_story_gen.csv
\begin{tabular}{lccc}
\toprule
 & High diversity & Low diversity & Gap \\
\midrule
$a_1$ (bits/byte) & 0.977 & 1.154 & +0.177 \\
$a_1$ (total bits) & 49.1 & 50.5 & +1.4 \\
Mean response bytes & 52.7 & 46.5 & $-6.3$ \\
$n$ & 103 & 97 & \\
\bottomrule
\end{tabular}

%% file: results/tables/confound_length.tex
% Generated by scripts/dataset_confound_analysis.py
% Source: results/tevet/qwen25_completion_v3/McDiv_nuggets/*_with_hds_story_gen.csv
\begin{tabular}{lccl}
\toprule
Byte range & High div.\ mean & Low div.\ mean & Gap \\
\midrule
$[20, 40)$ & 1.172 & 1.340 & +0.167 \\
$[40, 60)$ & 0.878 & 1.019 & +0.141 \\
$[60, 80)$ & 0.853 & 1.029 & +0.176 \\
$[80, 120)$ & 0.905 & 0.908 & +0.003 \\
\bottomrule
\end{tabular}

%% file: sections/appD_aggregation.tex
\section{Aggregation Across Prompts}\label{app:aggregation}

This appendix concerns aggregation across prompts.
Aggregation \emph{within} a single prompt, across permutations of the response ordering, is described in Section~\ref{sec:ordering} and given by Eq.~\eqref{eq:akbar-mor}: the per-byte $\bar{a}_k$ is a mean of per-permutation per-byte rates, not a ratio of permutation-averaged bits to permutation-averaged byte counts.\footnote{An earlier revision of our implementation computed the ratio-of-means form, in which the bits and byte counts at each position are each averaged across permutations before dividing.
After enough permutations, the byte counts at each position approach the mean response length, so that estimator collapses into the total-bits curve rescaled by a single constant, losing the per-permutation per-byte signal that Eq.~\eqref{eq:akbar-mor} preserves.
Switching the implementation to the form in Eq.~\eqref{eq:akbar-mor} shifts the headline numbers in this paper by 0--17\% (in mean-of-ratios' favor), with no qualitative change to any ranking we report.}

The diversity score $D_{Ca_n}$ is defined relative to a single prompt $p$.
To summarize a policy's diversity across a prompt suite $P = \{p_1, \ldots, p_M\}$, one can average:
\begin{equation}
    \bar{D}_{Ca_n}(\pi) = \frac{1}{M}\sum_{j=1}^{M} D_{Ca_n}(\pi, p_j)
\end{equation}

To compare two policies $\pi_1$ and $\pi_2$, the natural scalar is the difference:
\begin{equation}
    \Delta D_{Ca_n} = \bar{D}_{Ca_n}(\pi_1) - \bar{D}_{Ca_n}(\pi_2)
\end{equation}

$\Delta D_{Ca_n}$ answers ``how much more plausibility-weighted residual diversity does $\pi_1$ have than $\pi_2$, on average across prompts?''
No division is involved, so $\Delta D_{Ca_n}$ is stable even when either policy's score is near zero.

%% file: sections/appE_qwen3_comparison.tex
\section{Scaling the Base Model: Qwen2.5-3B vs Qwen3-30B-A3B-Base}\label{app:qwen3-comparison}

To test whether a stronger base model improves the metric on Tevet's benchmarks, we re-ran the full evaluation using Qwen3-30B-A3B-Base \citep{yang2025qwen3} (a 30B-parameter mixture-of-experts model with $\sim$3B active parameters per token) via the Tinker API,\footnote{\url{https://thinkingmachines.ai/tinker}} with the same setup as in Section~\ref{sec:tevet}: completion-format prompting, 50 permutations, no fine-tuning.
Table~\ref{tab:qwen3-comparison} reports per-task $C \times a_n$ (per-byte) for both base models.

\begin{table*}[!t]
\centering
\caption{Per-task comparison of $C \times a_n$ (per-byte) for Qwen2.5-3B vs Qwen3-30B-A3B-Base on Tevet diversity-eval. $\Delta$AUC is positive when the larger model wins.
ROC AUC is reported for the binary tasks (McDiv, McDiv\_nuggets, ConTest); only Spearman $\rho$ is meaningful for DecTest (continuous temperature labels).}
\label{tab:qwen3-comparison}
\small
\input{results/tables/qwen3_comparison.tex}
\end{table*}

\paragraph{Result: scaling did not help on the binary tasks.} On the \qwenThreeBinaryTotal{} binary classification tasks (McDiv, McDiv\_nuggets, ConTest), Qwen2.5-3B wins on \qwenThreeBinaryWinsQwenTwoFive{} of \qwenThreeBinaryTotal{}, with ConTest prompt\_gen the only marginal win for Qwen3-30B-A3B ($\qwenThreeConTestPromptGenDeltaAUC$ AUC).
The mean degradation from scaling is small but consistent (mean $\Delta$AUC = $\qwenThreeBinaryMeanDeltaAUC$ across the \qwenThreeBinaryTotal{} binary tasks).
On the \qwenThreeDecTestTotal{} DecTest tasks (continuous temperature labels), Qwen3-30B-A3B is mildly better on \qwenThreeDecTestWins{} of \qwenThreeDecTestTotal{} (mean $\qwenThreeDecTestMeanDeltaRho$ in Spearman $\rho$, with one tie).

\paragraph{Interpretation.} The binary result is a clear negative: scaling from 3B to 30B active parameters does not improve discrimination on McDiv or ConTest.
We do not have a confident explanation.
The Qwen3-30B-A3B run was not repeated after a bug fix applied to the primary experiments, so the comparison should be treated as preliminary.
We leave the question of whether larger base models improve the metric on binary tasks to future work.

%% file: results/tables/qwen3_comparison.tex
% Generated by: scripts/compare_qwen_models.py --emit-latex
% Data sources: figures/tevet_validation/c_ainf_analysis_v3/summary_table.txt
%               figures/tevet_validation/c_ainf_analysis_qwen3/summary_table.txt
% Metric: C x a_n (per-byte). Q2.5-3B = Qwen2.5-3B local; Q3-30B = Qwen3-30B-A3B-Base via Tinker.
\begin{tabular}{@{}llrrrrr@{}}
\toprule
 &  & \multicolumn{2}{c}{Qwen2.5-3B} & \multicolumn{2}{c}{Qwen3-30B-A3B} & \\
\cmidrule(lr){3-4} \cmidrule(lr){5-6}
Section & Task & $\rho$ & AUC & $\rho$ & AUC & $\Delta$AUC \\
\midrule
McDiv & prompt\_gen (no\_hds) & 0.729 & 0.921 & 0.718 & 0.915 & -0.006 \\
 & resp\_gen (no\_hds) & 0.500 & 0.788 & 0.483 & 0.779 & -0.009 \\
 & story\_gen (no\_hds) & 0.523 & 0.802 & 0.493 & 0.785 & -0.017 \\
\midrule
McDiv\_nuggets & prompt\_gen (no\_hds) & 0.636 & 0.867 & 0.615 & 0.855 & -0.012 \\
 & prompt\_gen (with\_hds) & 0.683 & 0.895 & 0.655 & 0.878 & -0.017 \\
 & resp\_gen (no\_hds) & 0.345 & 0.699 & 0.325 & 0.687 & -0.012 \\
 & resp\_gen (with\_hds) & 0.437 & 0.753 & 0.425 & 0.746 & -0.007 \\
 & story\_gen (no\_hds) & 0.317 & 0.683 & 0.293 & 0.669 & -0.014 \\
 & story\_gen (with\_hds) & 0.405 & 0.734 & 0.376 & 0.717 & -0.017 \\
\midrule
ConTest & prompt\_gen (with\_hds) & 0.584 & 0.837 & 0.592 & 0.842 & +0.005 \\
 & resp\_gen (with\_hds) & 0.391 & 0.726 & 0.333 & 0.692 & -0.034 \\
 & story\_gen (with\_hds) & 0.686 & 0.896 & 0.653 & 0.877 & -0.019 \\
\midrule
DecTest & prompt\_gen (no\_hds) & 0.842 & --- & 0.877 & --- & --- \\
 & prompt\_gen (with\_hds) & 0.845 & --- & 0.875 & --- & --- \\
 & resp\_gen (no\_hds) & 0.771 & --- & 0.804 & --- & --- \\
 & resp\_gen (with\_hds) & 0.760 & --- & 0.777 & --- & --- \\
 & story\_gen (no\_hds) & 0.763 & --- & 0.771 & --- & --- \\
 & story\_gen (with\_hds) & 0.785 & --- & 0.785 & --- & --- \\
\bottomrule
\end{tabular}

%% file: sections/appF_rlhf_cross_metric.tex
\section{Cross-Metric Agreement on the OLMo-2-7B RLHF Experiment}\label{app:rlhf-cross-metric}

The RLHF case study (Section~\ref{sec:rlhf-experiment}) reports the headline monotone $D_{Ca_n}$ drop in Table~\ref{tab:rlhf-diversity}.
\ifworkshop
Figure~\ref{fig:rlhf-metric-scatter} plots $D_{Ca_n}$ against the lexical (EAD) and semantic (SentBERT) baselines, confirming that the three metrics see the same diversity-loss signal as the $\bar{a}_k$ curves and per-prompt distributions in §\ref{sec:rlhf-experiment}.
\else
Figure~\ref{fig:rlhf-ak-violin} below visualises the underlying $\bar{a}_k$ curves and per-prompt $D_{Ca_n}$ distributions for the same length-matched subset, and Figure~\ref{fig:rlhf-metric-scatter} plots $D_{Ca_n}$ against the lexical (EAD) and semantic (SentBERT) baselines, confirming that the three metrics see the same diversity-loss signal.

\IfFileExists{figures/rlhf_experiment/ak_curves_overlay_alpacaeval_lm.pdf}{%
\begin{figure*}[h]
\centering
\begin{subfigure}[t]{0.48\textwidth}
    \includegraphics[width=\textwidth]{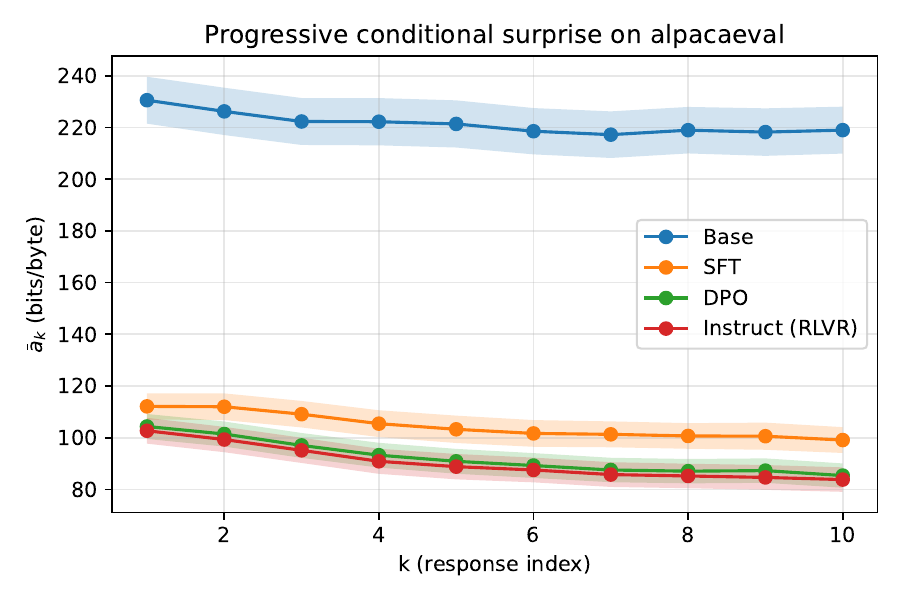}
    \caption{AlpacaEval $\bar{a}_k$ curves (length-matched).}
\end{subfigure}\hfill
\begin{subfigure}[t]{0.48\textwidth}
    \includegraphics[width=\textwidth]{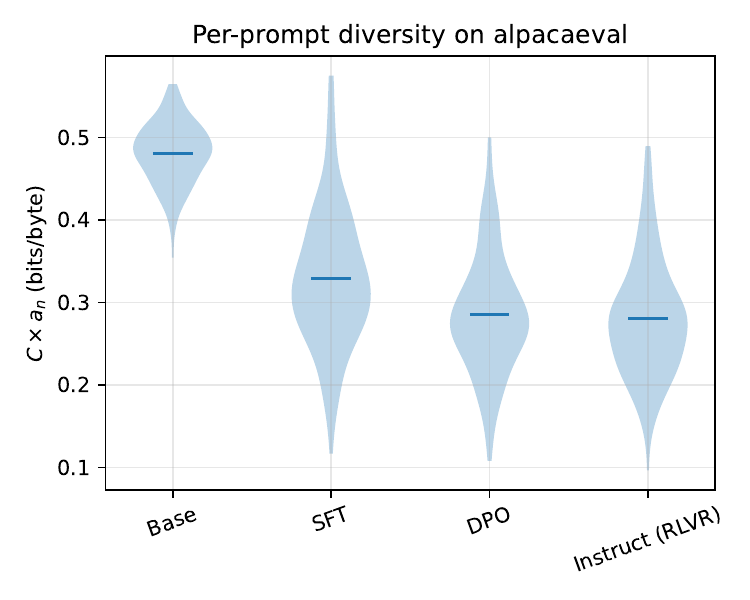}
    \caption{Per-prompt $D_{Ca_n}$ by stage (length-matched).}
\end{subfigure}
\caption{Progressive conditional surprise curves and per-prompt diversity distributions across the four OLMo-2-7B stages on the length-matched AlpacaEval subset. Each later stage's $\bar{a}_k$ curve lies below the base curve at every $k \geq 2$; the per-prompt $D_{Ca_n}$ distribution shifts toward lower values as the pipeline advances.}
\label{fig:rlhf-ak-violin}
\end{figure*}
}{}
\fi

\IfFileExists{figures/rlhf_experiment/metric_correlation_scatter_alpacaeval_lm.pdf}{%
\begin{figure*}[!htbp]
\centering
\begin{subfigure}[t]{\textwidth}
    \centering
    \includegraphics[width=0.85\textwidth]{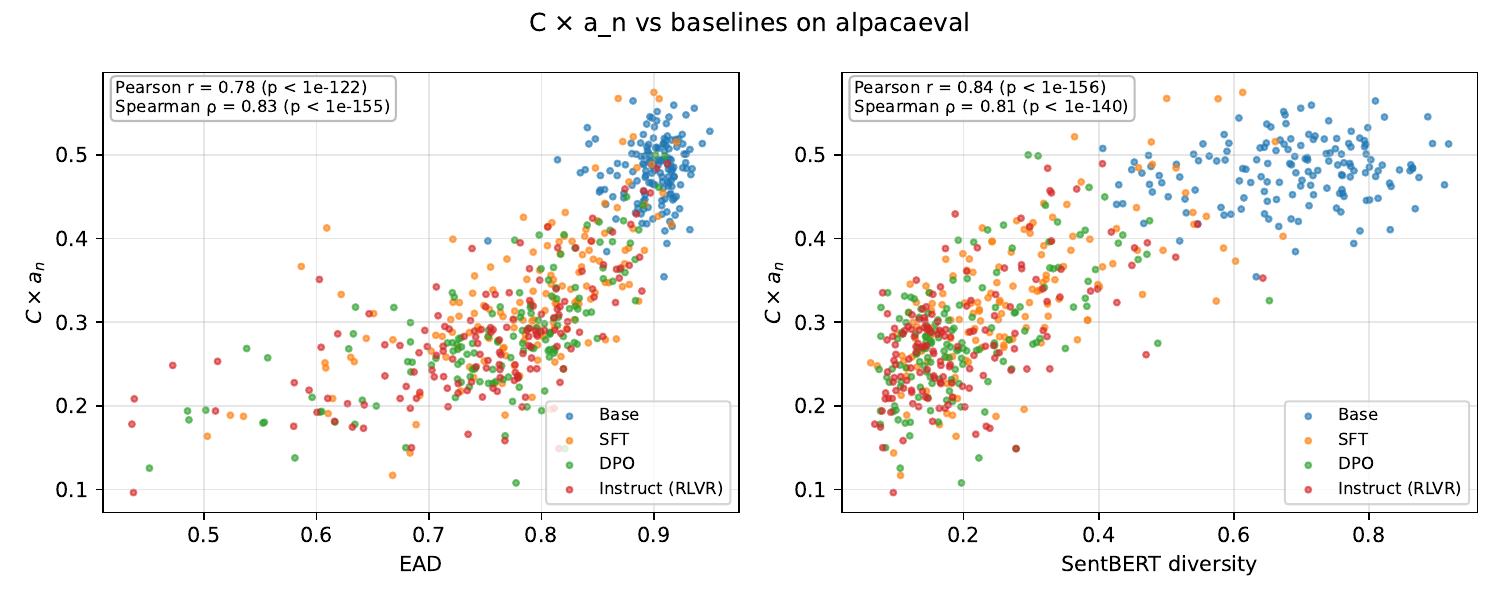}
    \caption{AlpacaEval (length-matched, $n=\olmoLmAlpacaN$ prompts).}
\end{subfigure}\\[0.5em]
\begin{subfigure}[t]{\textwidth}
    \centering
    \includegraphics[width=0.85\textwidth]{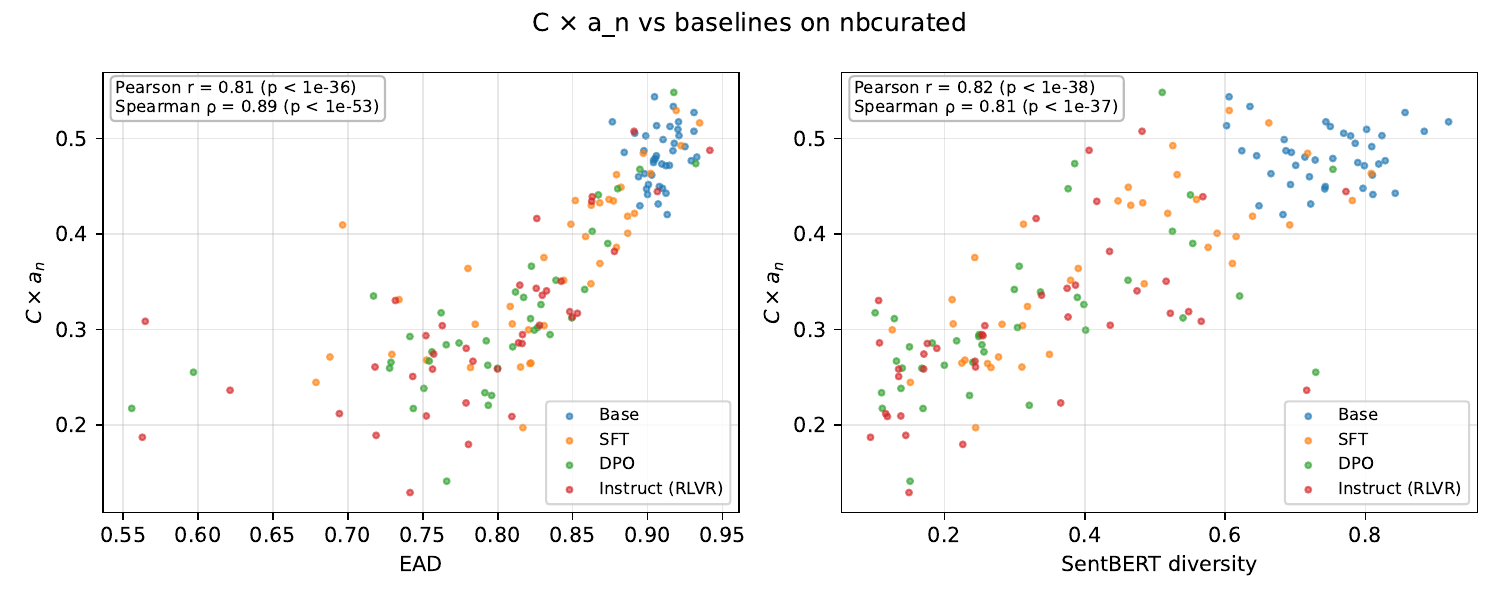}
    \caption{NoveltyBench curated (length-matched, $n=\olmoLmNbCuratedN$ prompts).}
\end{subfigure}
\caption{Per-prompt $D_{Ca_n}$ versus EAD (left subpanel) and SentBERT-similarity diversity (right subpanel), coloured by stage, on the length-matched subset of prompts. Length-matching truncates each (stage, prompt) tuple's responses to a common per-prompt byte budget so the per-byte conditional surprise that defines $a_n$ is not depressed by response length. Pearson $r$ and Spearman $\rho$ (two-sided $p$) are reported in each panel. The rank correlations are positive across both prompt sets and both baselines: $D_{Ca_n}$ tracks the same diversity-loss signal EAD and SentBERT detect, and the agreement is not an artefact of response length.}
\label{fig:rlhf-metric-scatter}
\end{figure*}
}{}